%% file: main.tex
\documentclass[runningheads]{llncs}

\usepackage{eccv}

\usepackage{eccvabbrv}

\usepackage{graphicx}
\usepackage{subcaption}
\usepackage{booktabs}
\usepackage{multirow}
\usepackage{makecell}
\usepackage{wrapfig}
\usepackage{algorithm}
\usepackage{algpseudocode}
\usepackage{marvosym}

\usepackage[accsupp]{axessibility}  % Improves PDF readability for those with disabilities.

\usepackage[pagebackref,breaklinks,colorlinks,citecolor=eccvblue]{hyperref}
\usepackage{orcidlink}

\makeatletter
\renewcommand*{\@fnsymbol}[1]{\ensuremath{\ifcase#1\or *\or \dagger\or \ddagger\or
   \mathsection\or \mathparagraph\or \|\or **\or \dagger\dagger
   \or \ddagger\ddagger \else\@ctrerr\fi}}
\makeatother

\let\titleold\title
\renewcommand{\title}[1]{%
\titleold{#1}
\newcommand{\thetitle}{#1}
}

\def\maketitlesupplementary{
{
\newpage
\centering
\Large
\vspace{0.5em}
\textbf{Appendix}
\vspace{1.0em}
}
}

\begin{document}

\title{Measuring 3D Spatial Geometric Consistency in Dynamic Video Generation} 

\titlerunning{SGC}

\author{
Weijia Dou\inst{1,2,}\thanks{\!\!Equal contributions. $^\dagger$Project leader. \textsuperscript{\Letter}Corresponding author.} \and
Wenzhao Zheng\inst{1,3,*,\dagger} \and
Weiliang Chen\inst{1} \and
Yu Zheng\inst{1} \and \\
Jie Zhou\inst{1} \and
Jiwen Lu\inst{1,\text{\Letter}}
}

\authorrunning{W. Dou, W. Zheng et al.}

\institute{
$^1$Tsinghua University \samelineand
$^2$Tongji University \samelineand
$^3$University of California, Berkeley\\
Code: \url{https://github.com/tj12323/SGC}\\
\texttt{renrendwj@tongji.edu.cn; wenzhao.zheng@outlook.com;\\lujiwen@tsinghua.edu.cn}
}
\newcommand{\samelineand}{\quad}

\maketitle

\input{sec/0_abstract}    
\input{sec/1_intro}

\input{sec/2_relatedworks}

\input{sec/3_method}
\input{sec/4_experiment}

\section*{Acknowledgements}
This work was supported in part by the National Natural Science Foundation of China under Grant 62441616, Grant 62321005, Grant 62125603, and Grant 62336004.

\bibliographystyle{splncs04}
\bibliography{main}

\input{sec/X_suppl}

\end{document}

%% file: sec/0_abstract.tex
\begin{abstract}
Recent generative models can produce high-fidelity videos, yet they often exhibit 3D spatial geometric inconsistencies. 
Existing evaluation methods fail to accurately characterize these inconsistencies: fidelity-centric metrics like FVD are insensitive to geometric distortions, while consistency-focused benchmarks often penalize valid foreground dynamics. 
To address this gap, we introduce SGC, a metric for evaluating 3D \textbf{S}patial \textbf{G}eometric \textbf{C}onsistency in dynamically generated videos. 
We quantify geometric consistency by measuring the divergence among multiple camera poses estimated from distinct local regions. Our approach first separates static from dynamic regions, then partitions the static background into spatially coherent sub-regions. 
We predict depth for each pixel, estimate a local camera pose for each subregion, and compute the divergence among these poses to quantify geometric consistency. 
Experiments on real and generative videos demonstrate that SGC robustly quantifies geometric inconsistencies, effectively identifying critical failures missed by existing metrics. 
\keywords{Video Generation \and Geometric Consistency}
\end{abstract}

%% file: sec/1_intro.tex
\section{Introduction}
\label{sec:intro}

\begin{figure*}[t]
    \centering
    \includegraphics[width=1\linewidth]{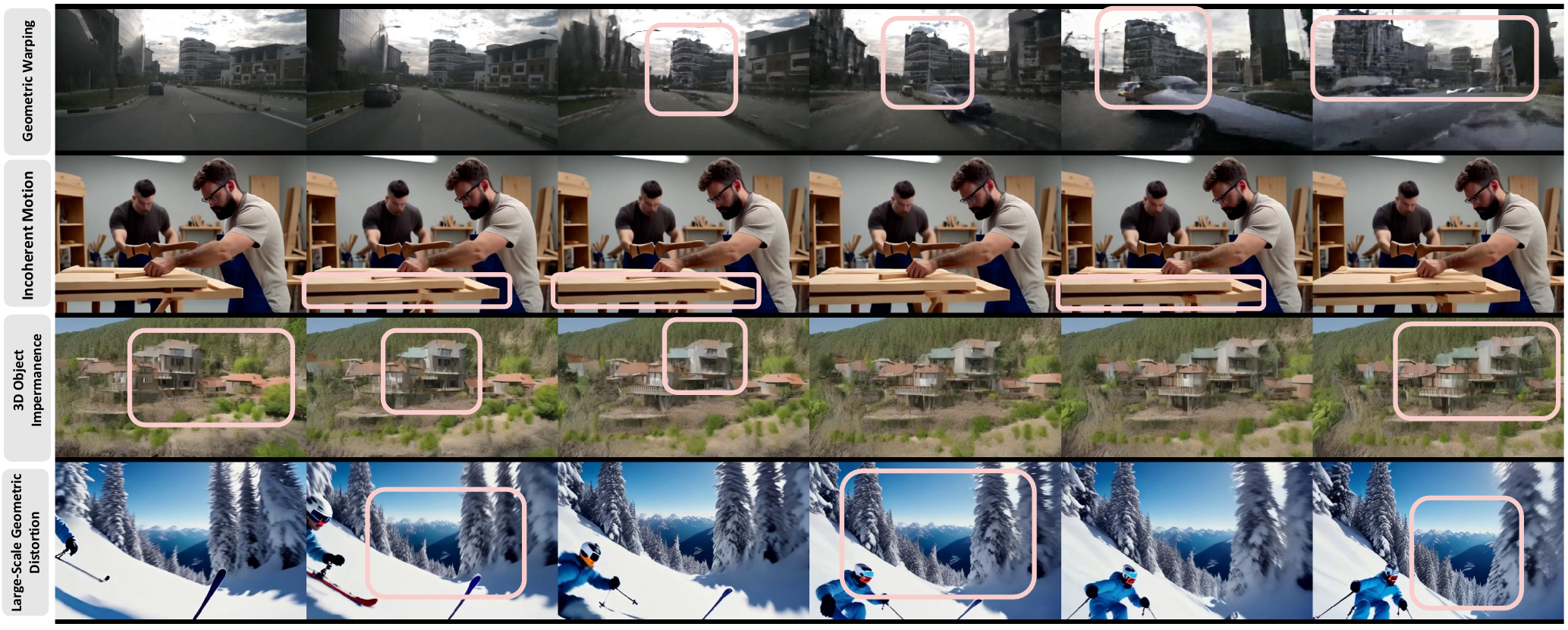}  
    \caption{\textbf{Examples of 3D Spatial Geometric Inconsistencies in Generated Videos.} Existing models often fail to maintain geometric consistency, exhibiting critical 3D spatial failures despite plausible per-frame visuals. (a) Geometric Warping: The rigid structure of the static buildings severely distorts as the camera moves. (b) Incoherent Motion: The static workbench illogically ``sticks'' to and moves with the dynamic piece of wood, violating physical separation. (c) Object Impermanence: A static structure on the mountain ``flickers'' and illogically changes its shape, failing to persist over time. (d) Perspective Failure: The distant mountains, which should remain stable, unnaturally warp and ``narrow'' as the skier moves forward, violating 3D perspective.}
    \vspace{-7mm}
    \label{fig:inconsistencies}
\end{figure*}

Recent generative video synthesis models achieve high visual fidelity~\cite{ma2024latte, bar2024lumiere, agarwal2025cosmos}, approaching photorealism and enabling content creation and simulation applications. 
Despite strong performance on fidelity-centric metrics like FVD ~\cite{unterthiner2018towards}, current models often exhibit significant 3D spatial geometric inconsistencies.  
As illustrated in \cref{fig:inconsistencies}, these failures include (1) geometric warping—static backgrounds distorting during camera motion, (2) incoherent motion—static and dynamic objects illogically fusing, (3) object impermanence—structures flickering or morphing across frames, and (4) perspective failures—large-scale scene distortions violating 3D perspective principles. 

A critical impediment to addressing these failures is the lack of robust metrics to measure 3D spatial geometric consistency in dynamically generated videos.
Existing metrics fall into two categories, each with fundamental limitations.
Fidelity-centric metrics (e.g., FVD~\cite{unterthiner2018towards}) exhibit a content-over-motion bias, i.e., they are dominated by per-frame appearance and are largely insensitive to geometric distortions, thus rewarding plausible textures despite unstable geometry~\cite{allen2025direct}.
Conversely, existing consistency-focused metrics suffer from a fragility-to-motion bias, i.e., metrics from novel view synthesis~\cite{watson2022novel, yu2023long} are not robust to dynamic objects, while benchmarks like VBench~\cite{huang2024vbench} may penalize valid vigorous motion~\cite{liu2024fr}.
Other specialized tools like FVMD~\cite{liu2024fr} focus on object kinematics while neglecting background stability.
This fragmentation highlights a critical gap: no established metrics effectively isolate and diagnose 3D spatial geometric consistency amidst complex foreground dynamics.

To address this gap, we propose SGC (\textbf{S}patial \textbf{G}eometric \textbf{C}onsistency), a diagnostic metric to quantify this foundational 3D spatial geometric consistency, explicitly focusing on the stability of the static background. 
SGC is founded on the physical principle that in a coherent 3D world, the apparent motion of all static background points ($V_{P_i} = 0$) must be consistent with a single, shared camera transformation ($T_{cam}$). 
Our method first isolates the static background to analyze scene stability. 
Leveraging depth information, 3D points within these static regions are reconstructed and segmented into spatially coherent sub-regions. 
For each sub-region, the relative camera pose is estimated via Perspective-n-Point (PnP) using 2D feature tracks and their corresponding 3D points. 
In a physically coherent scene, these local pose estimates must agree, and their divergence signals geometric inconsistency. 
The SGC score quantifies this geometric instability by aggregating the divergence among these local pose estimates (measured as inter-region variance), their agreement with a global camera trajectory, and additional criteria such as cross-frame depth alignment. 
We validate SGC on diverse videos from state-of-the-art generative models (spanning text-to-video, image-to-video, and video-to-video paradigms) and real-world sequences. Experiments confirm that SGC robustly quantifies these geometric inconsistencies, effectively identifying failures overlooked by existing metrics and establishing it as a necessary complementary diagnostic tool.

%% file: sec/2_relatedworks.tex
\section{Related Work}
\textbf{Video Quality Assessment.}
Existing video quality metrics are either insensitive to geometric failures or overly sensitive to valid motion. Fidelity-centric metrics, including Fréchet Video Distance (FVD)~\cite{unterthiner2018towards,ge2024content,luo2024beyond}, suffer from content-over-motion bias by rewarding plausible per-frame appearance while ignoring geometric distortions~\cite{allen2025direct}. Similarly, prompt-alignment metrics like CLIPScore~\cite{hessel2021clipscore} assess frame-level semantics, neglecting dynamic properties like temporal consistency. These metrics are ill-suited for diagnosing foundational geometric failures. Conversely, consistency-focused metrics exhibit fragility-to-motion bias. Novel View Synthesis metrics (e.g., TSED~\cite{yu2023long}, MEt3R~\cite{asim24met3r}) fail when dynamic objects are present. Broad benchmarks like VBench~\cite{huang2024vbench} provide holistic assessments; however, using their consistency dimensions as geometric stability proxies risks confounding by valid high dynamics. Specialized tools fall short: FVMD~\cite{liu2024fr} evaluates object kinematics while ignoring background stability. VideoPhy~\cite{bansal2024videophy} assesses broad physical plausibility but cannot isolate background consistency from foreground dynamics. Furthermore, while a recent study~\cite{li2024sora} analyzes Sora's geometric consistency via 3D reconstruction quality, an automated, model-agnostic metric for arbitrary dynamic videos remains lacking. As existing metrics can hardly diagnose background geometric stability amidst complex foreground motion, we introduce a physically grounded metric that explicitly disentangles background stability from foreground dynamics.

\noindent\textbf{Generative Video Models.}
Recent advancements in generative video models have led to distinct categories based on input modality, including Text-to-Video (T2V)~\cite{villegas2022phenaki,zhang2023controlvideo,wang2025lavie,chen2024videocrafter2,wang2023modelscope,ma2024latte,kondratyuk2023videopoet,bar2024lumiere,zheng2024open,kong2024hunyuanvideo,blattmann2023align,ge2023preserve,singer2022make,yuan2024inflation,jin2024pyramidal}, Image-to-Video (I2V)~\cite{zhang2024moonshot,xing2024dynamicrafter,zeng2024make,ni2024ti2v,chen2023seine,ren2024consisti2v,zhang2023i2vgen,blattmann2023stable,jin2024pyramidal}, and Video-to-Video (V2V)~\cite{gu2025das,agarwal2025cosmos}.
Architectural innovations like the Space-Time U-Net in Lumiere~\cite{bar2024lumiere} target global temporal consistency by generating the entire video in one pass. To expand input flexibility, VideoPoet~\cite{kondratyuk2023videopoet} leverages an LLM for zero-shot generation from text, images, and audio. 
Other works focus on high-fidelity synthesis and control~\cite{skorokhodov2024hierarchical,wang2025lingen,li2025cdragchainofthoughtdrivenmotion,geng2025motion,burgert2025go}. 
I2VGen-XL~\cite{zhang2023i2vgen} employs a cascaded diffusion strategy to separate semantic coherence from spatio-temporal refinement, while Diffusion as Shader~\cite{gu2025das} provides 3D-aware control via 3D tracking signals, enabling versatile manipulations. Models like Cosmos~\cite{agarwal2025cosmos} scale to model complex 3D environmental dynamics.

%% file: sec/3_method.tex
\section{Proposed Approach}

\begin{figure*}[t]
    \centering
    \includegraphics[width=1\linewidth]{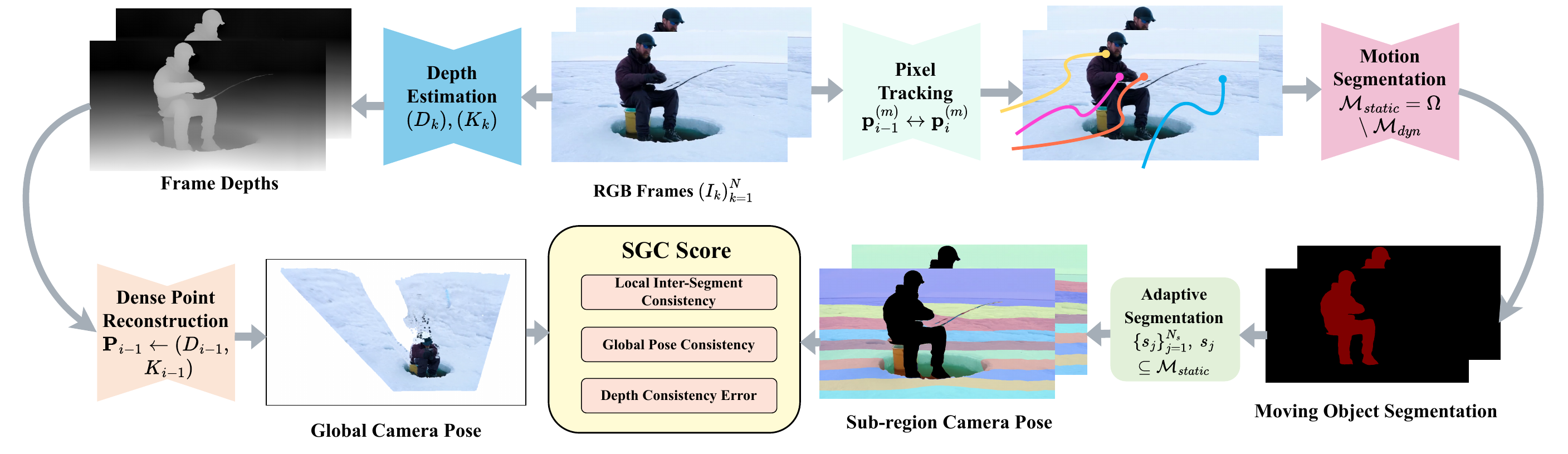}
    \caption{\textbf{Overview of the SGC computation pipeline.} Input RGB frames undergo parallel processing: (i) depth estimation, leading to dense point reconstruction for global camera pose estimation; and (ii) pixel tracking followed by motion segmentation to isolate moving objects. The identified static background is then adaptively segmented. Local camera poses for these static sub-regions are subsequently estimated using information from pixel tracks and depth. Finally, the overall SGC score is computed by aggregating three key evaluations: local inter-segment consistency, global pose consistency, and cross-frame depth consistency.}
    \vspace{-5mm}
    \label{fig:overview}
\end{figure*}

We introduce the $SGC$ score, a diagnostic metric designed to evaluate 3D spatial geometric consistency in dynamically generated videos. $SGC$ is premised on the physical principle that all points belonging to the static background must adhere to a single, coherent camera transformation. The metric quantifies violations of this principle by measuring the variance among multiple local camera pose estimates derived exclusively from these static regions.

The implementation follows the feedforward pipeline shown in \cref{fig:overview}. To directly address the ``fragility-to-motion'' bias discussed earlier, the pipeline begins by isolating the static background $\mathcal{M}_{static}$ (\cref{sec:moving-object-segmentation}). Subsequent steps then estimate global camera parameters and depth (\cref{sec:global-pose-intrinsic}), partition $\mathcal{M}_{static}$ into coherent sub-regions using depth (\cref{sec:local-relative-pose}), estimate local relative poses $P_{i}^{loc,j}$ for these sub-regions via PnP, and finally aggregate metrics quantifying the divergence among these local estimates and their agreement with the global motion $P_{i}^{glo}$ into the final $SGC$ score (\cref{sec:composite-evaluation-score}).

\subsection{Static Background Isolation}\label{sec:moving-object-segmentation}
A critical limitation of existing metrics is their ``fragility-to-motion'' bias, as they are often confounded by complex foreground dynamics. To establish a robust diagnostic for geometric consistency, our method first disentangles the static background from independently moving objects. Failure to do so introduces two key confounding factors: (1) a generator may be penalized for valid complex object motion, and (2) severe object-level inconsistencies can obscure the evaluation of an otherwise stable background.
Therefore, our metric is explicitly designed to diagnose the stability of the static background, complementing kinematic metrics that evaluate dynamic elements. To achieve this isolation, we employ SegAnyMo~\cite{huang2025segmentmotionvideos} as a preprocessing step. This method segments dynamic content by classifying long-range spatio-temporal point tracks, yielding precise per-frame moving-object masks. The union of these masks forms the dynamic region, denoted as $\mathcal{M}_{dyn}$. The static background region $\mathcal{M}_{static}$, which is the exclusive focus of our subsequent geometric analysis, is defined as the set complement:
\begin{equation}
\mathcal{M}_{static} = \Omega \setminus \mathcal{M}_{dyn}, 
\end{equation}
where $\Omega$ denotes the entire image spatial domain.

\subsection{Dense Point Reconstruction}\label{sec:global-pose-intrinsic}
To accurately reconstruct the 3D scene, SGC relies on two complementary geometry sources: global camera geometry and per-frame scene depth. 

Traditional Structure from Motion (SfM) pipelines (e.g., COLMAP~\cite{schonberger2016structure}) often fail in reconstructing 3D scenes from generative videos. To address these limitations, we consider the Visual Geometry Grounded Transformer (VGGT)~\cite{wang2025vggt}, a feed-forward neural network architected to directly infer a comprehensive suite of 3D scene attributes from a sequence of $N$ input images $(I_k)_{k=1}^{N}$, where $I_k \in \mathbb{R}^{3 \times H \times W}$. The model defines a function $f$ that maps these images to their corresponding 3D annotations:
\begin{equation}
f\left((I_k)_{k=1}^{N}\right) = \left( (R_{wc,k}, t_{wc,k}, K_k), D_k, P_k, T_k \right)_{k=1}^{N},
\label{eq:vggt_func_unified_narrative}
\end{equation}
where $(R_{wc,k}, t_{wc,k})$ represents the global camera pose (with $R_{wc,k} \in SO(3)$ being the world-to-camera rotation matrix and $t_{wc,k} \in \mathbb{R}^3$ the translation vector), $K_k \in \mathbb{R}^{3 \times 3}$ is the camera intrinsic matrix, $D_k$ is the depth map, $P_k$ is the viewpoint-invariant point map, and $T_k$ are dense features. 

Our system specifically exploits VGGT strictly for extracting robust global camera poses $(R_{wc,k}, t_{wc,k})$ and intrinsics $K_k$. Separately, to obtain high-quality and temporally consistent scene depth, we substitute VGGT depth output with Video Depth Anything~\cite{video_depth_anything} to estimate the per-frame dense depth map.

\subsection{Local Relative Pose Estimation}\label{sec:local-relative-pose}

We partition this region $\mathcal{M}_{static}$ into depth-consistent sub-regions.
We apply clustering to the 1D depth values $d_p$ associated with pixels $(x,y)$ within $\mathcal{M}_{static}$. 
We robustly isolate distinct 3D surfaces by first partitioning $\mathcal{M}_{static}$ into globally coherent depth strata (e.g., robustly isolating a distant mountain range from a mid-ground building) and subsequently treating the resulting depth-based groups as sub-regions for local estimation.

The set of $N_p$ valid depth values, $D = \{d_p \mid (x_p, y_p) \in \mathcal{M}_{static} \text{ and } d_p \text{ is valid} \}$, is processed by depth clustering into $k_{seg}$ clusters of depth magnitudes, $\{S_j\}$. Each cluster $S_j$ groups pixels from $\mathcal{M}_{static}$ that share similar depth values. These pixel groupings, after filtering by a minimum total count, constitute the $N_s$ ($N_s \le k_{seg}$) distinct subregions $s_j \subseteq \mathcal{M}_{static}$. Each sub-region $s_j$ in frame $f_i$ serves as an independent unit for subsequent pose estimation.
For each such subregion $s_j$ identified in frame $f_i$, we estimate its local relative camera pose with respect to the preceding frame $f_{i-1}$. This pose, denoted by the rotation $R_{i}^{loc,j} \in SO(3)$ and translation $\mathbf{t}_{i}^{loc,j} \in \mathbb{R}^3$, describes the transformation of points from the camera coordinate system of $f_{i-1}$ to that of $f_i$. 

Dense 2D tracking establishes the robust frame-to-frame correspondences required for local pose estimation. This yields per-pixel motion trajectories across consecutive frames, providing necessary 2D matching pairs. Specifically, for a set of 2D points $\{ \mathbf{p}_{i-1}^{(m)} = (u_{i-1}^{(m)}, v_{i-1}^{(m)}) \}$ observed in $f_{i-1}$ and their corresponding tracked locations $\{ \mathbf{p}_i^{(m)} = (u_i^{(m)}, v_i^{(m)}) \}$ falling within subregion $s_j$ in $f_i$, we first reconstruct their 3D coordinates in the camera frame of $f_{i-1}$. Given the depth map $D_{i-1}$ and the camera intrinsic matrix $K_{i-1}$ (obtained from the global geometry estimator), each 3D point $\mathbf{P}_{i-1}^{(m)}$ is reconstructed by unprojecting its 2D coordinates using the corresponding depth value and focal lengths. 

With the set of 3D points $\{\mathbf{P}_{i-1}^{(m)}\}$ and their 2D projections $\{\mathbf{p}_i^{(m)}\}$ in subregion $s_j$, along with the intrinsic matrix $K_i$ for frame $f_i$, the Perspective-n-Point (PnP) algorithm is employed to solve for $R_{i}^{loc,j}$ and $\mathbf{t}_{i}^{loc,j}$ that satisfy the projection equation for each correspondence $m$.

\subsection{$SGC$: Spatial Geometric Consistency Score}\label{sec:composite-evaluation-score}
To evaluate the 3D geometric consistency between consecutive frames ($f_{i-1}, f_i$), we define three component metrics that leverage the $N_s$ local relative poses $P_{i}^{loc,j} = (R_{i}^{loc,j}, \mathbf{t}_{i}^{loc,j})$ estimated from static sub-regions and the global relative pose $P_{i}^{glo} = (R_{i}^{glo}, \mathbf{t}_{i}^{glo})$. High errors indicate failures to render a coherent 3D scene evolving consistently over time.

\begin{figure}[t]
    \centering
    \vspace{-5mm}
    \includegraphics[width=1\linewidth]{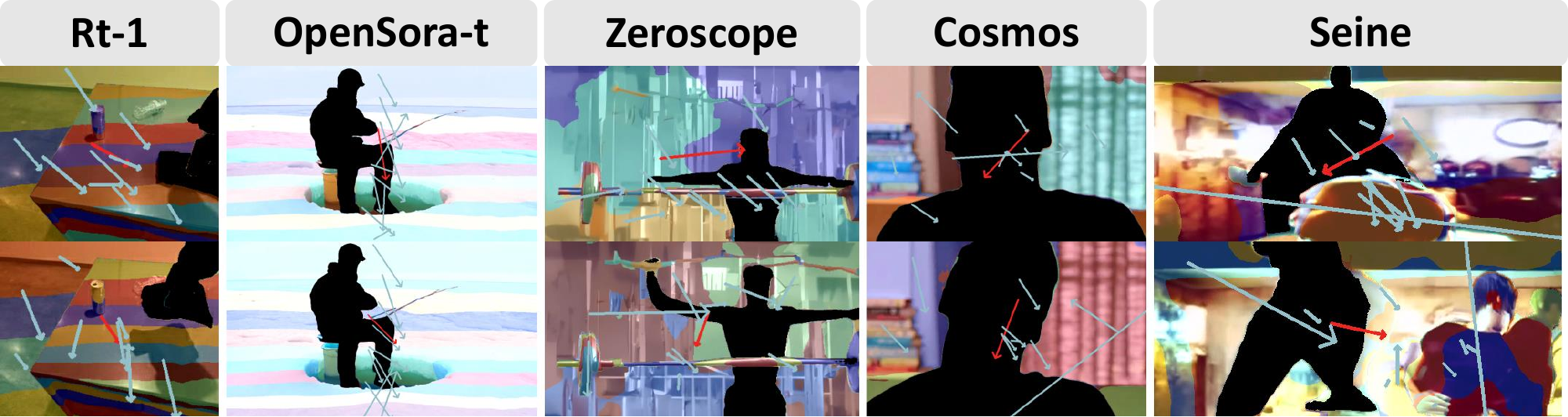}    
    \caption{\textbf{Visualization of the Local Inter-Segment and Global Pose Consistency.}     
    Blue arrows depict image-plane projections of motion directions induced by local relative poses $P_{i}^{loc,j}$ of static sub-regions, while the red arrow denotes the projected global relative pose $P_{i}^{glo}$. Arrows are obtained by projecting a unit 3D direction vector under each relative transformation, length-normalized for visualization. Sequences from left to right exhibit decreasing consistency scores. (Left) High consistency: local poses are similar and align with global motion. (Right) Low consistency: local poses show high variance and/or deviate significantly from global motion.}
    \vspace{-5mm}
    \label{fig:pose_viz}
\end{figure}

\textbf{Local Inter-Segment Consistency.} This metric measures motion estimation stability within the static background. Disagreement among local poses $\{P_{i}^{loc,j}\}$ from different static sub-regions indicates the generator's inability to maintain a rigid scene structure.
We quantify this via two variance measures:

$\bullet$ Local Rotational Variance:
\begin{equation}
\sigma_{rot,loc}^{2} = \frac{1}{N_s} \sum_{j=1}^{N_s} (d_{\theta}(R_{i}^{loc,j}, \overline{R}_{i}^{loc}))^{2},
\end{equation}
where $d_{\theta}(R_A, R_B) = \arccos\left(\frac{\text{Trace}(R_A^T R_B) - 1}{2}\right)$ and $\bar{R}_{i}^{loc}$ is the mean local rotation. 

$\bullet$ Local Translational Variance:
\begin{equation}
\sigma_{trans,loc}^2 = \frac{1}{N_s} \sum_{j=1}^{N_s} \bigl\| \mathbf{t}_{i}^{loc,j} - \bar{\mathbf{t}}_{i}^{loc} \bigr\|_2^2,
\end{equation}
where $\bar{\mathbf{t}}_{i}^{loc}$ is the mean local translation.

\textbf{Global Pose Consistency.} This metric assesses the alignment between locally estimated motions and the global pose $P_{i}^{glo}$. Significant deviations suggest that local dynamics are erratic or disconnected from the plausible global camera movement, breaking large-scale spatio-temporal coherence (see \cref{fig:pose_viz}). We evaluate this using:

$\bullet$ Global Rotational Variance:
\begin{equation}
\sigma_{rot,glob}^{2} = \frac{1}{N_s} \sum_{j=1}^{N_s} (d_{\theta}(R_{i}^{loc,j}, R_{i}^{glo}))^{2}.
\end{equation}

$\bullet$ Global Translational Variance:
\begin{equation}
\sigma_{trans,glob}^2 = \frac{1}{N_s} \sum_{j=1}^{N_s} \bigl\| \mathbf{t}_{i}^{loc,j} - \mathbf{t}_{i}^{glo} \bigr\|_2^2.
\end{equation}

\textbf{Depth Consistency Error ($E_{depth}$).} This metric evaluates the stability of the 3D scene geometry over time using the global pose $P_{i}^{glo}$ for alignment. It measures the discrepancy between the depth map of $f_i$ ($D_i$) and the warped depth map from $f_{i-1}$ ($D_{i-1 \to i}$). A large error suggests the video fails to maintain a consistent 3D world.
\begin{equation}
E_{depth} = \frac{1}{|\mathcal{V}|} \sum_{\mathbf{u} \in \mathcal{V}} \bigl| D_{i-1 \to i}(\mathbf{u}) - D_i(\mathbf{u}) \bigr|.
\end{equation}
where $\mathcal{V}$ is the set of valid overlapping pixels in the static scene mask of $f_i$.

\textbf{Overall $SGC$ Score.} 
To aggregate these $N_M$ component metrics (where $N_M$ denotes the number of component metrics) into a single, robust $SGC$ score, we employ an objective weighting procedure. First, all raw component scores $M_{d,k}$ are normalized to a common $[0, 1]$ scale (where 0 is best). This normalization, involving Z-score standardization followed by min-max scaling, is crucial to prevent metrics with high raw magnitudes (e.g., translational variance) from dominating the final score. Second, to avoid subjective manual tuning, we derive fixed weights $w_k$ by performing Principal Component Analysis (PCA) on the normalized score matrix from a diverse calibration dataset. The weights $w_k$ are the normalized loadings of the first principal component (PC1). This process automatically assigns greater influence to metrics capturing the most significant shared variance related to geometric inconsistency. The final $SGC$ score for a video $d$ is the weighted sum of its normalized component scores $M''_{d,k}$:
\begin{equation}
SGC_d = \sum_{k=1}^{N_M} w_k M''_{d,k}.
\label{eq:final_SGC}
\end{equation}
This multi-component, variance-based design ensures $SGC$ is robust to camera motion magnitude, penalizing geometric inconsistency rather than the motion.

%% file: sec/4_experiment.tex
\section{Experiments}\label{sec:exps}
\subsection{Experimental Setup}\label{sec:details}

\noindent\textbf{Datasets and Models.}
To rigorously evaluate spatial geometric consistency (SGC), we curate a comprehensive benchmark of 1,296 videos. This comprises 1,196 videos from the GenWorld dataset~\cite{chen2025genworlddetectingaigeneratedrealworld} alongside an additional 100 high-motion web videos sampled from OpenVidHD-0.4M. To establish robust ground-truth anchors for real-world geometry and dynamics, we utilize 300 real videos across three distinct domains: \textbf{nuScenes}~\cite{caesar2020nuscenes} provides complex, multi-object driving scenarios; \textbf{RT-1}~\cite{brohan2022rt} offers highly dynamic robotic interactions; and \textbf{OpenVid}~\cite{nan2024openvid} serves as a crucial anchor that significantly enhances the diversity and representativeness of everyday web captures. The AI-generated portion consists of 996 videos sourced entirely from GenWorld, synthesized by 10 state-of-the-art generative models. To ensure comprehensive evaluation across paradigms, these models are grouped into: (1) \textbf{Text-to-Video (T2V)}~\cite{ma2024latte,zeroscope2024,wang2023modelscope,chen2024videocrafter2,hotshot2023,wang2025lavie,zheng2024open}; (2) \textbf{Image-to-Video (I2V)}~\cite{chen2023seine,zheng2024open}; and (3) \textbf{Video-to-Video (V2V)}~\cite{agarwal2025cosmos}. Detailed dataset statistics are deferred to the App.~\ref{sec:dataset}.

\noindent\textbf{Baselines.}
We compare our method against established metrics, categorized as follows: (1)\textbf{Feature/Consistency:} MEt3R~\cite{asim24met3r} and its motion-segmented variant, MEt3R(+MOS); (2)\textbf{Benchmark Suites:} VBench~\cite{huang2024vbench} (specifically Background Consistency, Dynamic Degree, and a weighted composite); and (3) \textbf{Motion/Physics Proxies:} FVD~\cite{unterthiner2018towards}, FVMD~\cite{liu2024fr}, and TRAJAN~\cite{allen2025direct}. Moving object segmentation (MOS) is exclusively applied to MEt3R to create the MEt3R(+MOS) baseline, whereas all other baselines operate on unmasked videos. Further baseline implementation details are provided in the App.~\ref{sec:baseline}.

\noindent\textbf{Unified Evaluation Protocol.}
To ensure a strictly fair comparison, all evaluations adhere to a unified protocol. To isolate geometric stability, we define a static background mask, $\mathcal{M}_{static}$, where masked foreground pixels are explicitly ignored. For MEt3R(+MOS), this exact masking is applied by filtering the score map prior to final calculation. Furthermore, we standardize the underlying foundational estimators: VGGT~\cite{wang2025vggt} for global poses and intrinsics, Video Depth Anything~\cite{video_depth_anything} for depth, DELTA~\cite{ngo2024delta} for dense tracking, and SegAnyMo~\cite{huang2025segmentmotionvideos} for MOS extraction. To evaluate local consistency, we robustly isolate distinct 3D surfaces by applying a GPU-accelerated $k$-means clustering to the 1D depth values $d_p$ associated with valid pixels $(x,y)$ within $\mathcal{M}_{static}$. Finally, to aggregate SGC components, PCA weights are learned exactly once on the full dataset. Additionally, the VBench weighted composite metrics are calculated using this identical weighting method on our dataset to ensure calibration parity. Detailed hyperparameter specifications are provided in the App.~\ref{sec:hyper}.

\subsection{Benchmarking Geometric Consistency in Generative Models}

\begin{table}[t]
  \centering
  \small
  \caption{\textbf{Quantitative comparison of generative video methods and real-world video datasets in terms of 3D spatial geometric consistency.} The table presents an evaluation of the performance using MEt3R~\cite{asim24met3r}, FVD~\cite{unterthiner2018towards}, the proposed SGC metric, and VBench~\cite{huang2024vbench} dimensions: background consistency (BC), dynamic degree (DD), weighted score (w), and TRAJAN~\cite{allen2025direct} and FVMD~\cite{liu2024fr}. The best and second-best scores are highlighted in \textbf{bold} and \underline{underlined}, respectively.}
  \label{tab:genworld-vertical}
\vspace{-3mm}
  \setlength{\tabcolsep}{2pt}
  \resizebox{\textwidth}{!}{
  \begin{tabular}{@{}c|c|c|ccccccccc@{}}
  \toprule
  \multirow{2}{*}{} 
    & \multirow{2}{*}{\centering\textbf{Methods}}
    & \multirow{2}{*}{\centering\textbf{Task}}
    & \multirow{2}{*}{\textbf{SGC ↓}}
    & \multirow{2}{*}{\textbf{MEt3R~\cite{asim24met3r} ↓}}
    & \multirow{2}{*}{\makecell{\textbf{MEt3R~\cite{asim24met3r}} \\ \textbf{(+MOS) ↓}}}
    & \multirow{2}{*}{\textbf{FVD~\cite{unterthiner2018towards} ↓}}
    & \multicolumn{3}{c}{\textbf{VBench~\cite{huang2024vbench} ↑}}
    & \multirow{2}{*}{\textbf{TRAJAN~\cite{allen2025direct} ↑}}
    & \multirow{2}{*}{\textbf{FVMD~\cite{liu2024fr} ↓}} \\
  & & & & & & 
    & \textbf{BC} & \textbf{DD} & \textbf{weighted} & \\
  \midrule

  \multirow{10}{*}{\textbf{Gen.}}
    & Cosmos~\cite{agarwal2025cosmos} & V2V
    & 0.0722 & 0.0741 & 0.2350 & 760.3204
    & 0.9461 & 0.6462 & 0.6037 & 0.6023 & 15389.47 \\
    & Hotshot~\cite{hotshot2023} & T2V
    & 0.1172 & 0.1371 & 0.3568 & 908.0547
    & 0.9468 & 0.8190 & 0.6801 & 0.5518 & - \\
    & Latte~\cite{ma2024latte} & T2V
    & 0.3226 & 0.1707 & 0.2601 & 801.0069
    & 0.9412 & 0.8559 & 0.4403 & 0.4418 & 19561.51 \\
    & Lavie~\cite{wang2025lavie} & T2V
    & 0.1241 & 0.1122 & 0.2793 & 679.8203
    & \underline{0.9576} & 0.7559 & 0.7090 & \underline{0.6504} & \underline{14424.44} \\
    & Modelscope~\cite{wang2023modelscope} & T2V
    & 0.3129 & 0.1851 & 0.4208 & 784.6681
    & 0.9196 & 0.7966 & 0.3313 & 0.5962 & 19511.57 \\
    & OpenSora-i~\cite{zheng2024open} & I2V
    & 0.1631 & 0.1030 & 0.3088 & \textbf{751.2137}
    & 0.8993 & 0.8333 & 0.3638 & 0.3208 & \textbf{9250.18} \\
    & OpenSora-t~\cite{zheng2024open} & T2V
    & 0.0831 & 0.0919 & 0.2513 & \underline{838.0382}
    & 0.9259 & 0.8286 & 0.5370 & 0.2835 & 13169.15 \\
    & Seine~\cite{chen2023seine} & I2V
    & 0.2837 & 0.2613 & 0.6343 & 808.9575
    & 0.8891 & \underline{0.8981} & 0.1215 & 0.4669 & 23555.54 \\
    & VideoCrafter~\cite{chen2024videocrafter2} & T2V
    & 0.0973 & 0.1176 & 0.2205 & 699.7098
    & \textbf{0.9650} & 0.6320 & \underline{0.7114} & 0.6344 & 16125.86 \\
    & Zeroscope~\cite{zeroscope2024} & T2V
    & 0.0912 & 0.1558 & 0.4602 & 823.1359
    & 0.9326 & 0.5287 & 0.4974 & 0.5230 & 18714.92 \\

  \midrule

  \multirow{3}{*}{\textbf{Real}}
    & OpenVid~\cite{nan2024openvid} & --
    & \textbf{0.0530} & \textbf{0.0371} & \textbf{0.0486} & 752.2600
    & 0.9408 & 0.6300 & \textbf{0.8330} & 0.4869 & - \\
    & RT-1~\cite{brohan2022rt} & --
    & 0.0639 & 0.0799 & \underline{0.0698} & 1352.8758
    & 0.9229 & \textbf{0.9800} & 0.6499 & \textbf{0.7362} & - \\
    & nuScenes~\cite{caesar2020nuscenes} & --
    & \underline{0.0613} & \underline{0.0485} & 0.0753 & 1297.0720
    & 0.9080 & 0.8800 & 0.5370 & 0.2320 & - \\

  \bottomrule
  \end{tabular}  
  }
  \vspace{-7mm}
\end{table}

\noindent\textbf{Unified Benchmark and Overall Performance.}
We present a comprehensive quantitative evaluation in Table 1, benchmarking 10 contemporary generative models against three diverse real-world anchors (RT-1, nuScenes, and OpenVid). To provide a rigorous and multi-faceted assessment, our unified benchmark compares the proposed SGC metric against established Feature/Consistency metrics (MEt3R~\cite{asim24met3r}, MEt3R(+MOS)), Benchmark Suites (VBench~\cite{huang2024vbench} BC, DD, and weighted composite), and Motion/Physics Proxies (FVD~\cite{unterthiner2018towards}, FVMD~\cite{liu2024fr}, and TRAJAN~\cite{allen2025direct}). As expected, the real-world datasets establish a strict lower bound for spatial error. Among the generative methods, Cosmos~\cite{agarwal2025cosmos} achieves the most competitive SGC score (0.0722), though it still exhibits a noticeable gap when compared to real-world anchors like OpenVid (0.0530) or nuScenes (0.0613). Conversely, models such as Seine~\cite{chen2023seine}, Modelscope~\cite{wang2023modelscope}, and Latte~\cite{ma2024latte} yield significantly higher SGC scores, indicating pronounced difficulties in maintaining spatial coherence across frames.

\begin{figure}[t]
    \centering
    \vspace{-5mm}
    \includegraphics[width=1\linewidth]{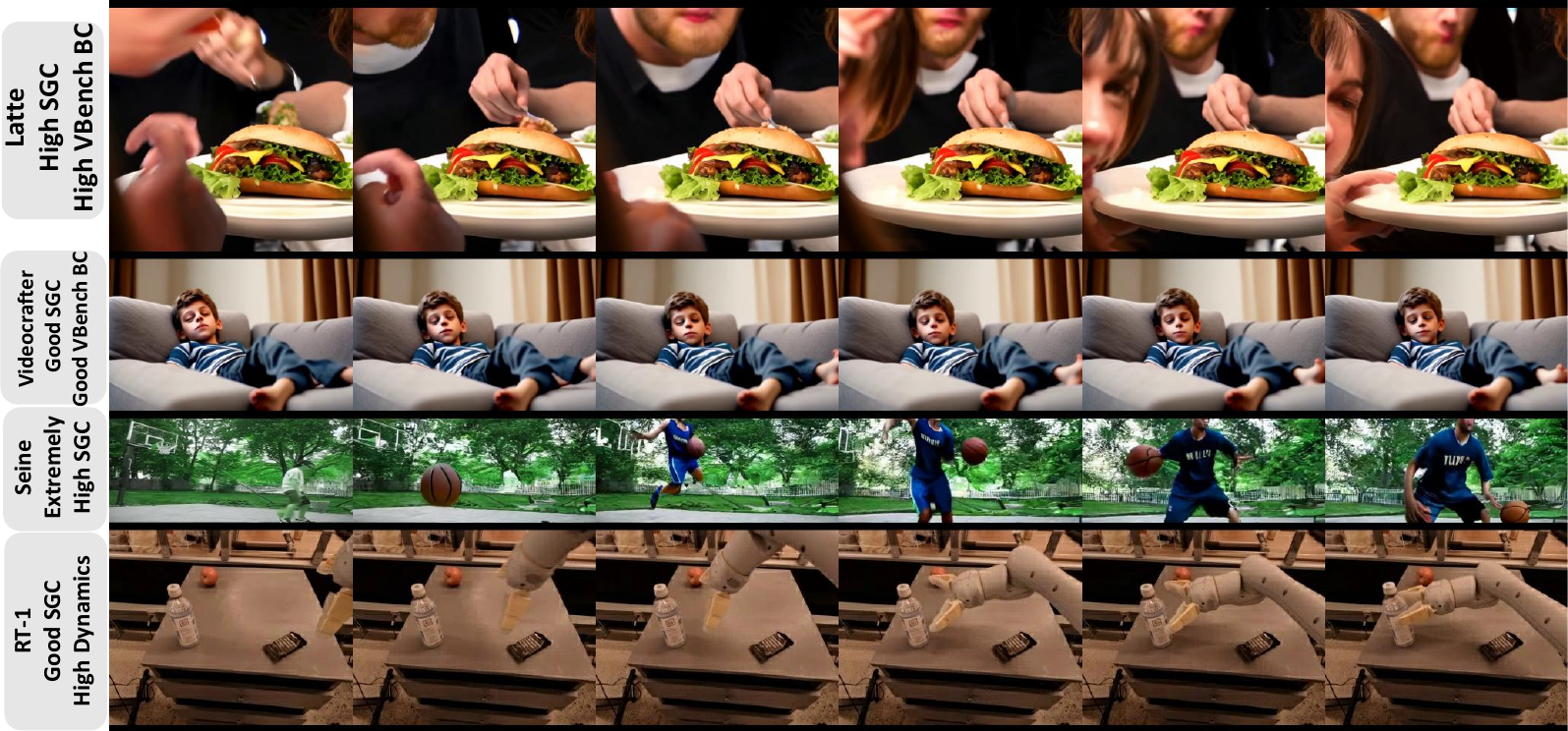}
\vspace{-7mm}
    \caption{\textbf{Qualitative Validation: SGC Detects Geometric Failures Missed by Feature Metrics.} Latte (R1) scores poorly on SGC (object instability), despite a high VBench-BC score from plausible textures. VideoCrafter (R2) is consistent, scoring well on both. Seine exhibits catastrophic geometric breakdown (e.g., subject distortion), reflected in its very high SGC score. RT-1, despite high dynamics (robot motion), scores excellently, demonstrating SGC motion robustness.}
    \vspace{-7mm}
    \label{fig:qualitative-show}
\end{figure}

\noindent\textbf{Isolating Background Dynamics.}
Comparing the standard MEt3R~\cite{asim24met3r} metric with its motion-segmented counterpart, MEt3R(+MOS), yields a critical observation. Across all evaluated generative models, explicitly filtering the score map using the static background mask $\mathcal{M}_{static}$ induces a distinct and consistent upward shift in error scores. Rather than representing a baseline failure, this adjustment isolates the background by stripping away foreground motion priors. This shift demonstrates unmasked feature-matching metrics frequently conflate accurate tracking of foreground textures with true underlying environmental stability. While MEt3R(+MOS) successfully prevents dynamic foregrounds from artificially inflating perceived consistency, both MEt3R variants remain fundamentally constrained to 2D feature tracking. Compared to SGC, MEt3R(+MOS) lacks an explicit understanding of scene depth and camera ego-motion. By leveraging foundational estimators like VGGT~\cite{wang2025vggt} and Video Depth Anything~\cite{video_depth_anything}, SGC evaluates true 3D background rigidity and multi-plane stability, providing a strictly more rigorous geometric constraint than 2D texture persistence alone. This distinction explains the performance gap for models like Cosmos~\cite{agarwal2025cosmos}. Although Cosmos achieves a competitive SGC score, it exhibits significantly higher error under MEt3R(+MOS). A component breakdown reveals Cosmos performs well within real-world ranges for depth consistency and translation, struggling primarily with rotation. While MEt3R penalizes 2D feature persistence failures, SGC correctly rewards this underlying 3D structural preservation.

\noindent\textbf{Diagnosing Foundational 3D Instabilities.}
The ranking derived from SGC diverges fundamentally from those produced by fidelity-centric and 2D physics metrics. For example, Latte and Modelscope achieve strong FVD~\cite{unterthiner2018towards} scores and high VBench~\cite{huang2024vbench} Background Consistency (BC) ratings, despite ranking poorly under SGC (as visualized in \cref{fig:qualitative-show}). This discrepancy illustrates that current generative models can synthesize highly plausible 2D textures (satisfying VBench-BC) and achieve high per-frame visual quality (satisfying FVD) while failing to preserve multi-plane stability and structural integrity. Ultimately, SGC is specifically sensitive to these foundational 3D geometric instabilities that remain hidden under texture-plausible generations, successfully exposing spatial inconsistencies that persist even after rigorous MOS adaptation.
Furthermore, a blinded pairwise human study demonstrating that SGC aligns with human judgments of static-background 3D stability is detailed in the App.~\ref{sec:userstudy}.

\subsection{Controlled Validation of Geometric Sensitivity}

\begin{figure}[t]
    \centering
    \vspace{-4mm}
    \includegraphics[width=0.85\linewidth]{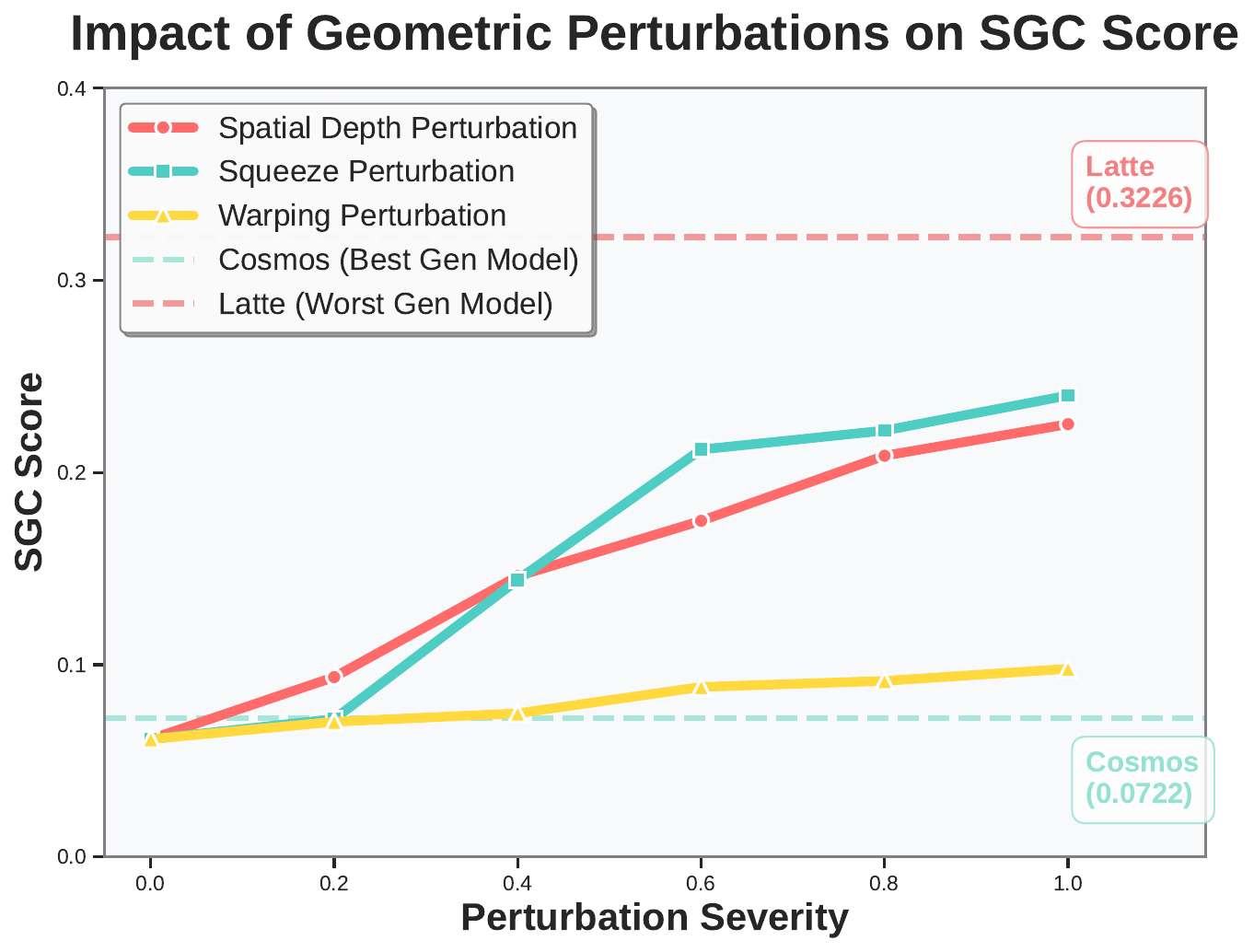}
    \vspace{-4mm}
    \caption{\textbf{Impact of Geometric Perturbations on SGC Score.} This line plot presents a sensitivity and monotonicity analysis, illustrating SGC scores rising with geometric degradation severity (from 0.0 to 1.0) on the nuScenes dataset. Baseline (unperturbed) SGC scores for the best-performing (Cosmos, 0.0722) and worst-performing (Latte, 0.3226) generative models on the GenWorld dataset are included as dashed teal and red reference lines, respectively. This controlled test confirms that high SGC scores directly measure geometric failure and not semantics or texture shifts.}
    \label{fig:sgc_perturbatio}
    \vspace{-7mm}
\end{figure}

\noindent\textbf{Synthetic Perturbations on Real Videos.}
To address potential domain shift criticisms and verify that our metric isolates true geometric integrity, we conduct a causal-style controlled validation. We systematically inject three types of synthetic perturbations into the high-fidelity nuScenes dataset. To accurately simulate the progressive structural collapse often observed in generative models, the perturbation intensity scales linearly with time, compounding the distortions in later frames. We evaluate three distinct corruption paradigms: (1) \textbf{Warping:} A time-varying, non-rigid sinusoidal deformation applied jointly to the RGB image and 2D feature tracks, simulating continuous spatial drift while holding depth estimations constant. (2) \textbf{Perspective Squeeze:} A global anisotropic vertical compression applied exclusively to the 2D tracks. By keeping the RGB image static, this creates a severe cross-modal conflict between the apparent visual scale and the underlying multi-view geometry. (3) \textbf{Spatial Depth Warp:} A synchronized spatial remapping applied simultaneously across the depth map, RGB image, and 2D tracks. This aggressively disrupts temporal consistency by inducing continuous depth boundary drift and non-rigid trajectory shifts.

\begin{figure}[t]
    \centering
    \includegraphics[width=1\linewidth]{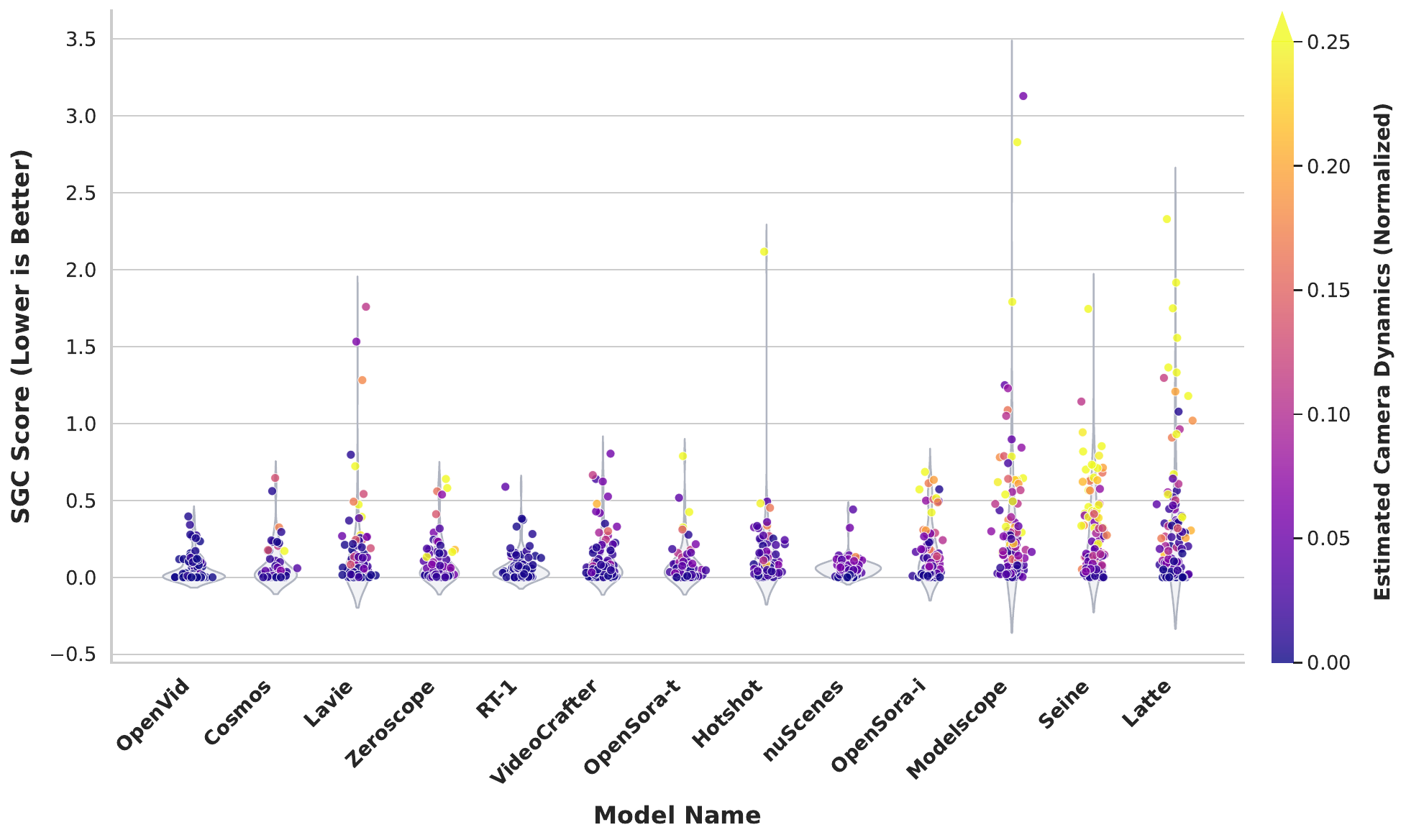}
    \vspace{-9mm}
    \caption{\textbf{Motion-Stratified SGC Distributions Across Models.}
    Violin plots with jittered samples showing the distribution of SGC scores (lower is better) for each model. Individual points are colored by the normalized Estimated Camera Dynamics (ECD).}
    \label{fig:stc_vs_ecd}
    \vspace{-6mm}
\end{figure}

\noindent\textbf{Sensitivity and Monotonicity.}
By strictly fixing the semantic content and texture quality, we evaluate SGC responsiveness purely to geometric degradation. As \cref{fig:sgc_perturbatio} illustrates, SGC exhibits a strict monotonic response to increasing perturbation severity (from 0.0 to 1.0) across all three corruption types. From a clean baseline score of 0.0613, the controlled injection of warping steadily inflates the SGC error to 0.0979. Spatial Depth Warp and Perspective Squeeze prove significantly more destructive to 3D consistency, precipitating steep degradation trajectories that culminate in SGC scores of 0.2252 and 0.2401, respectively.

The clear monotonic trend and strong rank correlation with perturbation severity empirically validate our core design. Ultimately, these controlled tests demonstrate that SGC reliably tracks underlying 3D spatial instabilities independent of visual appearance. This significantly reduces the likelihood that observed performance disparities between models are merely artifacts of semantics or texture shifts, proving SGC strictly isolates and measures geometric failure.

\subsection{Robustness, Disentanglement, and Diagnostic Analysis}

\noindent\textbf{Robustness to Camera Motion.} 
A reliable 3D consistency metric must decouple true geometric instability from the natural artifacts of camera movement. Existing metrics, such as the VBench Dynamic Degree (DD), conflate object dynamics with camera dynamics. To explicitly isolate camera motion, we introduce Estimated Camera Dynamics (ECD), derived from the global relative pose. We compute the average translational and rotational magnitudes per frame, min-max normalize them across the dataset, and define ECD as the maximum of these scaled values. On real-world datasets, SGC exhibits only weak to moderate correlations with ECD (RT-1: $r=0.44$, $p<0.001$; nuScenes: $r=0.52$, $p<0.001$; OpenVid: $r=0.17$, $p=0.09$). The corresponding coefficients of determination ($R^2$) indicate that camera dynamics explain at most 27\% of the SGC variance in RT-1 and nuScenes, and less than 3\% in OpenVid. This confirms that SGC is inherently robust to natural camera motion, with motion explaining only a minor fraction of error variation in real data.

\noindent\textbf{Disentangling Geometry from Motion.} 
To verify that SGC penalizes structural inconsistency rather than motion intensity, we conduct a motion-stratified visual analysis (\cref{fig:stc_vs_ecd}). While generated videos exhibit a stronger correlation between motion and SGC error (e.g., Latte: $r=0.79$, Modelscope: $r=0.47$), this sensitivity does not fully account for their elevated scores. Crucially, geometrically inconsistent models yield high SGC errors even under near-zero camera dynamics (dark-colored points), indicating a foundational deficit in maintaining 3D structure that is independent of motion. Conversely, real-world anchors securely maintain tightly bounded, low-SGC distributions (predominantly $<0.5$) even at the upper bounds of estimated camera motion (bright-colored points). This demonstrates that structural geometric collapse, rather than motion intensity, is the dominant factor driving SGC degradation in generative models.

\noindent\textbf{Complementarity to Fidelity Metrics.} 
By successfully isolating spatial instability from motion artifacts, SGC serves as a critical complement to standard fidelity metrics such as Fréchet Video Distance (FVD). Conventional metrics are heavily biased toward 2D texture plausibility and frame-level visual quality, frequently overlooking severe underlying structural degradation. The consistent separation of SGC scores between real and generated videos—irrespective of camera dynamics—highlights its unique capability to diagnose foundational 3D geometric failures that traditional metrics fail to capture.

\subsection{Ablation of SGC Design Choices}
\begin{table}[t]
\centering
\scriptsize
\setlength{\tabcolsep}{2pt}
\vspace{-5mm}
\caption{\textbf{Ablation study on SGC components across generative methods and real-world datasets.} We report SGC scores (↓) for three variants: without moving object segmentation (w/o MOS), grid-based depth clustering, and the full model (SGC-full). Method names correspond to cited works.}
\label{tab:ablation-stc-extended}
\vspace{-3mm}
\resizebox{\textwidth}{!}{
\begin{tabular}{l|cccccccccc|ccc}
\toprule
\textbf{Variant} 
& Cosmos
& Hotshot
& Latte
& Lavie
& Modelscope
& OpenSora-i
& OpenSora-t
& Seine
& VideoCrafter
& Zeroscope
& OpenVid
& RT-1
& nuScenes \\
\midrule

w/o MOS
& 0.052 & 0.104 & 0.286 & 0.133 & 0.222
& 0.157 & 0.080 & 0.259 & 0.086 & 0.075
& 0.014 & 0.163 & 0.064 \\

Grid depth
& 0.067 & 0.109 & 0.353 & 0.170 & 0.314
& 0.190 & 0.107 & 0.306 & 0.101 & 0.071
& 0.062 & 0.093 & 0.163 \\

Depth + Spatial 
& 0.082 & 0.133 & 0.555 & 0.161 & 0.483
& 0.205 & 0.101 & 0.365 & 0.086 & 0.084
& 0.019 & 0.051 & 0.153 \\

SGC-full
& 0.072 & 0.117 & 0.323 & 0.124 & 0.313
& 0.163 & 0.083 & 0.284 & 0.097 & 0.091
& 0.053 & 0.064 & 0.061 \\

\bottomrule
\end{tabular}
}
\end{table}

\begin{figure}[t]
    \centering
    \vspace{-4mm}
    \centering
    \begin{subfigure}[t]{0.24\linewidth}
        \centering
        \includegraphics[height=0.9\textwidth]{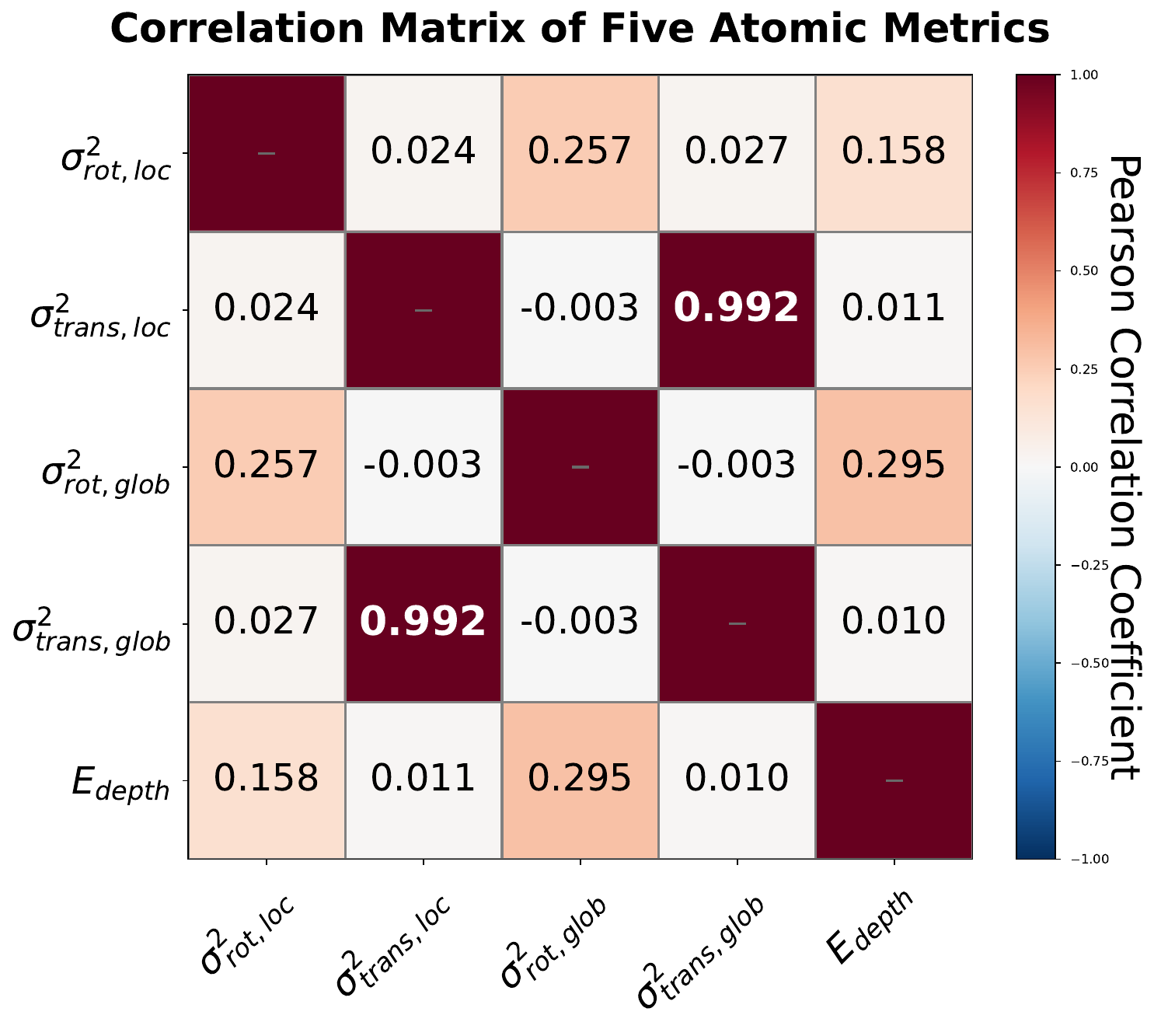}
        \caption{}
        \label{fig:sgc-component-analysis_a}
    \end{subfigure}
    \hfill
    \begin{subfigure}[t]{0.74\linewidth}
        \centering
        \includegraphics[height=0.28\textwidth]{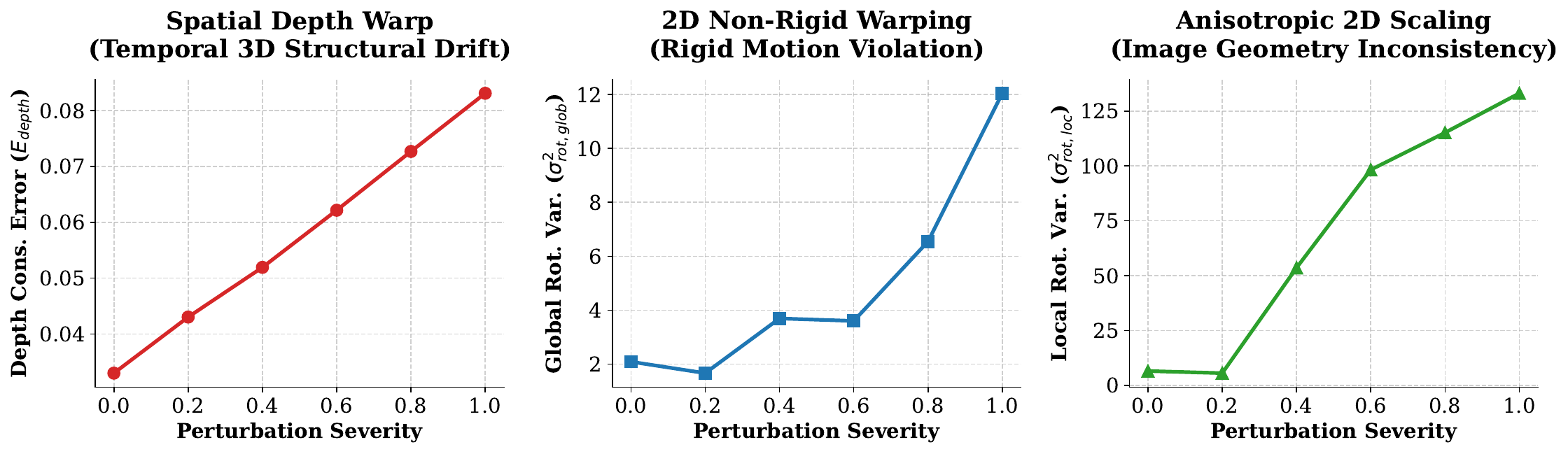}
        \caption{}
        \label{fig:sgc-component-analysis_b}
    \end{subfigure}
    \vspace{-3mm}
    \caption{\textbf{Validation of SGC components.} 
    Left: Pearson correlation matrix shows low inter-metric redundancy. 
    Right: Response curves for component-specific perturbations.}
    \vspace{-6mm}
    \label{fig:sgc-component-analysis}
\end{figure}

\noindent\textbf{Full-Dataset Architectural Ablations.}
We validate SGC core design choices by ablating individual modules across 10 generative models and 3 real-world anchors (\cref{tab:ablation-stc-extended}). First, we examine Moving Object Segmentation (MOS). Removing this module (w/o MOS) disrupts evaluating dynamic environments; on RT-1, the error increases from 0.064 to 0.163, as valid foreground articulation is erroneously penalized as background instability. Conversely, on largely static-camera datasets like OpenVid, removing MOS yields a lower error (0.014) than the full pipeline (0.053). This mild full-model increase occurs because segmentation in largely fixed-view scenes may introduce boundary noise or mask leakage, culling otherwise reliable background points and slightly weakening spatial constraints for pose estimation. Nonetheless, for in-the-wild videos with salient dynamic foregrounds, MOS remains important to prevent object motion from corrupting geometric evaluation. Further sensitivity analysis on how MOS quality impacts the SGC score is detailed in App.~\ref{sec:mossensitivity}.

Next, we analyze the spatial partitioning strategy. Replacing depth-aware plane clustering with a naive spatial grid (Grid depth) degrades performance in complex scenes (e.g., nuScenes error increases from 0.061 to 0.163), indicating grid-based partitioning fails to isolate distinct geometric structures. To further interrogate our 1D depth clustering, we evaluate a Depth + Spatial variant to address the hypothesis that clustering by depth alone might group non-contiguous surfaces and yield spatially insufficient regions for PnP estimation. While 1D depth clustering can merge disjoint surfaces (such as forming thin horizontal bands across flat roads), this behavior can be beneficial in practice. Aggregating non-contiguous but co-planar patches often increases the effective image-plane span of correspondences, typically improving the conditioning of PnP estimates (especially for orientation components). Enforcing spatial contiguity can over-segment expansive surfaces into compact sub-regions; these spatially restricted regions tend to yield less stable local pose estimates, inflating the real-world nuScenes error to 0.153 and increasing instability penalties across generative models (e.g., Latte increases from 0.323 to 0.555). Overall, these results support allowing 1D depth clustering to merge disjoint regions when doing so preserves a broad spatial distribution, improving robust geometric verification.

\noindent\textbf{Component Construct Validity.}
To verify SGC aggregated components capture distinct, interpretable failure modes, we analyze their empirical correlations and responses to controlled degradations. A Pearson correlation matrix across the dataset reveals most sub-metrics capture independent failure modes (\cref{fig:sgc-component-analysis_a}). While local ($\sigma_{trans,loc}^2$) and global ($\sigma_{trans,glob}^2$) translational variances exhibit high correlation ($r \approx 0.992$), this redundancy is an intentional architectural choice. This specific pair performs a hierarchical diagnostic check, comparing internal structural rigidity against global scene coherence, securely anchoring the PCA-weighted aggregation. To further validate this aggregation strategy, a cross-validation confirming the stability and generalizability across diverse video domains is detailed in the App.~\ref{sec:val_pca}.
Crucially, targeted synthetic perturbations verify individual SGC components directly map to their intended physical interpretations (\cref{fig:sgc-component-analysis_b}). Each controlled degradation explicitly isolated its corresponding metric: continuous warping exclusively inflated global pose divergence, driving global rotation variance from a baseline of 2.090 to 12.042. Applying the Spatial Depth Warp predictably triggered depth inconsistency, raising depth error from 0.033 to 0.083. Finally, simulating perspective collapse via the Squeeze perturbation induced severe cross-plane rigidity violations, causing local rotation variance to spike from 6.522 to 133.091. This diagnostic alignment confirms SGC is not merely a generalized error scalar but a composite of necessary components rigorously isolating and identifying specific 3D geometric failures.

\noindent\textbf{Estimator Robustness and Reliability.} A potential concern regarding SGC is its reliance on foundational estimators like VGGT, which are primarily trained on static scenes, for dynamic video evaluation. To validate that SGC reliably captures intrinsic geometric failures rather than estimator artifacts, we conducted a robustness analysis replacing VGGT with state-of-the-art dynamic-scene estimators (Any4D~\cite{karhade2026any4d} and Page4D~\cite{zhou2026paged}), alongside an ablation entirely removing the VGGT global pose component. 

As shown in \cref{tab:ranking_consistency}, substituting or removing the underlying estimator preserves remarkably strong rank correlations (Spearman $\rho \ge 0.860$). This confirms that the severe spatial inconsistencies detected in generative models are foundational structural failures, intrinsically captured by SGC within the evaluated estimators and datasets. Note that nuScenes was excluded from this specific subset analysis, as current dynamic-scene estimators like Any4D struggle to produce stable reconstructions in such complex, large-scale outdoor environments.

\begin{table}[t]
\centering
\caption{\textbf{Ranking consistency across geometric estimators.} NuScenes is excluded because Any4D~\cite{karhade2026any4d} fails on this outdoor dataset. All correlations are significant at $p<0.001$. Ranks per estimator are shown as superscripts.}
\vspace{-3mm}
\label{tab:ranking_consistency}
\resizebox{\textwidth}{!}{%
\footnotesize
\setlength{\tabcolsep}{2.5pt}
\begin{tabular}{lccccccccccccr}
\toprule
Estimator & OpenVid & RT-1 & Cosmos & OpenSora-t & Zeroscope & VideoCrafter & Hotshot & Lavie & OpenSora-i & Seine & Modelscope & Latte & $\rho$ vs Ori. \\
\midrule
Original  & 0.053\textsuperscript{1} & 0.064\textsuperscript{2} & 0.072\textsuperscript{3} & 0.083\textsuperscript{4} & 0.091\textsuperscript{5} & 0.097\textsuperscript{6} & 0.117\textsuperscript{7} & 0.124\textsuperscript{8} & 0.163\textsuperscript{9} & 0.284\textsuperscript{10} & 0.313\textsuperscript{11} & 0.323\textsuperscript{12} & 1.000 \\
w/o VGGT  & 0.077\textsuperscript{1} & 0.097\textsuperscript{2} & 0.107\textsuperscript{3} & 0.123\textsuperscript{5} & 0.122\textsuperscript{4} & 0.153\textsuperscript{6} & 0.183\textsuperscript{7} & 0.190\textsuperscript{8} & 0.198\textsuperscript{9} & 0.405\textsuperscript{10} & 0.460\textsuperscript{11} & 0.472\textsuperscript{12} & 0.993 \\
Any4D~\cite{karhade2026any4d} & 0.080\textsuperscript{2} & 0.062\textsuperscript{1} & 0.170\textsuperscript{3} & 0.226\textsuperscript{5} & 0.347\textsuperscript{9} & 0.224\textsuperscript{4} & 0.324\textsuperscript{8} & 0.322\textsuperscript{7} & 0.297\textsuperscript{6} & 0.836\textsuperscript{11} & 0.979\textsuperscript{12} & 0.809\textsuperscript{10} & 0.860 \\
Page4D~\cite{zhou2026paged} & 0.072\textsuperscript{2} & 0.062\textsuperscript{1} & 0.100\textsuperscript{5} & 0.091\textsuperscript{4} & 0.115\textsuperscript{6} & 0.091\textsuperscript{3} & 0.119\textsuperscript{7} & 0.121\textsuperscript{8} & 0.164\textsuperscript{9} & 0.310\textsuperscript{11} & 0.271\textsuperscript{10} & 0.332\textsuperscript{12} & 0.937 \\
\bottomrule
\end{tabular}
}
\vspace{-6mm}
\end{table}

\section{Conclusion}\label{sec:con}
We presented SGC, a novel diagnostic metric for 3D spatial geometric consistency, addressing a gap where existing fidelity (e.g., FVD) and consistency (e.g., VBench-BC, MEt3R) metrics fail. These metrics are often insensitive to geometric warping or are confounded by valid foreground motion, respectively. SGC overcomes this by employing a physically grounded principle: all static background points must adhere to a single camera transformation. It enforces this by isolating the static background $\mathcal{M}_{static}$ (avoiding penalties for valid motion) and measuring the variance among local pose estimates to quantify geometric integrity. Ablations confirmed isolating motion is essential for fairness, and we proved SGC is robust to camera motion magnitude. Using this validated metric, we find many SOTA models suffer from significant geometric failures missed by other metrics, despite high FVD or VBench scores. SGC thus serves as an essential, complementary diagnostic to guide research towards generating truly geometrically coherent videos.

%% file: sec/X_suppl.tex
\clearpage
\setcounter{page}{1}
\setcounter{table}{0}  
\maketitlesupplementary

\appendix

\section{Limitations and Future Work}
\label{sec:supp_limitations}

While SGC provides a robust, physically grounded metric for evaluating 3D spatial geometric consistency, it possesses certain practical constraints that present opportunities for future research.

\noindent\textbf{Scope of Static Backgrounds.} SGC is strictly founded on the physical constraint that static background points must align with a single, rigid camera transformation. By explicitly masking out dynamic foregrounds (via MOS) to prevent penalizing valid motion, assessing the internal, non-rigid 3D geometric consistency of moving entities (\eg, human articulation or fluid dynamics) falls outside our current scope.

\noindent\textbf{Estimator and Tracking Reliability.} The pipeline relies on off-the-shelf foundational estimators (\eg, VGGT, Video Depth Anything) and dense 2D tracking. Deploying models trained on physically grounded scenes to evaluate artifact-heavy generative videos introduces risks of domain shift and correspondence failures (\eg, due to extreme motion blur or morphing). While our empirical analysis demonstrates strong relative rank stability across different geometric backbones ($\rho \ge 0.860$), SGC currently lacks an absolute error calibration mechanism to entirely disentangle inherent estimator noise from generative geometric corruption during catastrophic scene breakdowns.

\noindent\textbf{Future Work.} Extending geometric consistency metrics to encompass complex non-rigid foreground dynamics—potentially by integrating articulated 3D motion priors or 4D reconstruction constraints—and developing estimator-agnostic absolute error bounds will be crucial next steps toward holistic, physics-based video evaluation.

\section{More Methodology and Algorithmic Details}
\subsection{Method Overview}
The complete SGC evaluation pipeline is detailed in ~\cref{algo:sgc_local,algo:sgc_aggregation}. Initially, global camera poses and dense depth maps are extracted using the VGGT model, allowing for the isolation of the static background, $\mathcal{M}_{static}$, via motion segmentation. This isolated static region is subsequently partitioned into coherent sub-regions through depth clustering. For each consecutive frame pair, local relative camera poses are estimated per sub-region by utilizing PnP-RANSAC on filtered 2D pixel tracks and their corresponding backprojected 3D coordinates. These local pose estimates enable the precise calculation of rotational and translational variances—evaluated against both the global trajectory and local means—alongside cross-frame depth errors. To prevent metrics with inherently large raw magnitudes from dominating the evaluation, the variance components undergo a logarithmic transformation, followed by rigorous dataset-level Z-score standardization and min-max normalization. Ultimately, the normalized metrics are aggregated into the final $SGC$ score using fixed weights derived from Principal Component Analysis, ensuring a robust, parameter-free quantification of 3D spatial geometric consistency.

\begin{algorithm*}[t]
\footnotesize
\setlength{\itemsep}{-1pt}
\caption{Local Metric Computation for SGC}
\label{algo:sgc_local}
\begin{algorithmic}[1]
    \Statex \textbf{Input:} RGB frames $(I_k)_{k=1}^{N}$, VGGT model $f$, depth clustering $k_{seg}$.
    \Statex \textbf{Output:} Per-pair metrics $\mathcal{R}$.
    
    \State Load $N$ RGB frames; Apply VGGT: $(R_{wc,k}, t_{wc,k}, K_k, D_k, P_k, T_k)_{k=1}^{N} \leftarrow f((I_k)_{k=1}^{N})$.
    \For {$k \leftarrow 1$ to $N$}
        \State Orthogonalize $R_{wc,k} \in SO(3)$.
    \EndFor
    \State $\mathcal{M}_{dyn} \leftarrow$ segment\_motion$(T_k)$; $\mathcal{M}_{static} \leftarrow \Omega \setminus \mathcal{M}_{dyn}$.
    \State Initialize $\mathcal{R} \leftarrow \{\}$.
    \For {$i \leftarrow 2$ to $N$}
        \State $\{s_j\}_{j=1}^{N_s} \leftarrow$ depth\_cluster\_and\_filter$(D_i, \mathcal{M}_{static}, k_{seg})$.
        \State $\{\mathbf{p}_{i-1}^{(m)}\}, \{\mathbf{p}_i^{(m)}\} \leftarrow$ load\_and\_filter\_tracks$(f_{i-1}, f_i)$.
        \For {$j \leftarrow 1$ to $N_s$}
            \State Filter tracks by subregion; Backproject to 3D; Estimate local pose via PnP-RANSAC.
        \EndFor
        \State Compute global pose: $P_{i}^{glo}$; Local means: $\bar{R}_{i}^{loc}, \bar{\mathbf{t}}_{i}^{loc}$.
        \State Compute variances: $\sigma_{rot,loc}^{2}, \sigma_{trans,loc}^2, \sigma_{rot,glob}^{2}, \sigma_{trans,glob}^2$.
        \State Compute depth error: $E_{depth}$.
        \State Append metrics to $\mathcal{R}$.
    \EndFor
    \State \textbf{return} $\mathcal{R}$.
\end{algorithmic}
\end{algorithm*}

\begin{algorithm*}
\footnotesize
\caption{SGC Score Aggregation with Dataset-Level Normalization}
\label{algo:sgc_aggregation}
\begin{algorithmic}[1]
    \Statex \textbf{Input:} Per-pair metrics $\mathcal{R}$ from ~\cref{algo:sgc_local}.
    \Statex \textbf{Output:} $SGC$ score.
    
    \Statex \textit{// Dataset-level parameter computation (performed once)}
    \State Compute per-metric averages: $\bar{M}_k \leftarrow \text{mean}(\mathcal{R}_k)$.
    \For {$k \in \{\sigma_{rot,loc}^{2}, \sigma_{trans,loc}^2, \sigma_{rot,glob}^{2}, \sigma_{trans,glob}^2\}$}
        \State $\bar{M}_k^{log} \leftarrow \log(1 + \bar{M}_k)$ \Comment{Log transform for variance metrics}
    \EndFor
    \State Compute Z-score parameters across dataset: $\mu_k \leftarrow \text{mean}(\bar{M}_k^{log})$, $\sigma_k \leftarrow \text{std}(\bar{M}_k^{log})$.
    \State Compute Z-scores: $Z_k \leftarrow (\bar{M}_k^{log} - \mu_k) / \sigma_k$.
    \State Compute Min-Max bounds: $z_{min,k} \leftarrow \min(Z_k)$, $z_{max,k} \leftarrow \max(Z_k)$.
    \State Perform PCA on normalized metrics to derive weights $(w_k)_{k=1}^{N_M}$ where $\sum w_k = 1$.
    
    \Statex
    \Statex \textit{// Video-level normalization using pre-computed parameters}
    \State Apply log transform to current video's variance metrics.
    \State Standardize: $Z_k \leftarrow (\bar{M}_k^{log} - \mu_k) / \sigma_k$ using dataset $\mu_k, \sigma_k$.
    \State Normalize and clip: $M''_k \leftarrow \max\left(0, \frac{Z_k - z_{min,k}}{z_{max,k} - z_{min,k}}\right)$.
    \State Compute weighted score: $SGC \leftarrow \sum_{k=1}^{N_M} w_k M''_k$.
    \State \textbf{return} $SGC$.
\end{algorithmic}
\end{algorithm*}

\subsection{Final PCA Component Weights}\label{sec:pca}
As stated in ~\cref{sec:composite-evaluation-score,sec:details} of the main paper, the final SGC score is an aggregation of its component metrics. To avoid subjective manual tuning, these components are weighted objectively using Principal Component Analysis (PCA) on the normalized score matrix from our 1196-video calibration dataset.

The weights are derived from the normalized loadings of the first principal component (PC1). This component captures the axis of maximum variance in the data, which, in this context, represents the most significant and shared mode of 3D geometric inconsistency across all evaluated videos. A component weight thus reflects its relative importance as an indicator of this primary failure mode.

The final, fixed weights ($w_k$) used to compute all $S_{3DC}$ scores in this paper are presented in ~\cref{tab:pca_weights}.

An analysis of the derived PCA loadings reveals a distinct trend: translational errors and depth inconsistencies capture the most significant shared variance. Specifically, global translational variance ($\sigma_{trans,glob}^2$, $w=0.2459$), local translational variance ($\sigma_{trans,loc}^2$, $w=0.2403$), and depth consistency error ($E_{depth}$, $w=0.2307$) heavily dominate the PC1 axis, collectively determining the majority of the metric's sensitivity. In contrast, rotational deviations ($\sigma_{rot,glob}^2$ and $\sigma_{rot,loc}^2$) exhibit comparatively lower relative weights of $0.1665$ and $0.1167$, respectively. This distribution objectively demonstrates that across current generative architectures, the primary degradation of 3D spatial geometric consistency predominantly manifests as severe translational structural drift and cross-frame depth misalignment rather than rotational instability, thus validating our composite design in accurately isolating these critical spatial failures.

\begin{table}[t]
\centering
\caption{Final PCA-Derived Component Weights ($w_k$).}
\label{tab:pca_weights}
\vspace{-3mm}
\begin{tabular}{cc}
\hline
Metric Component ($M_k$) & PCA Weight ($w_k$) \\
\hline
\textbf{$\sigma_{trans,glob}^2$} & \textbf{0.2459} \\
\textbf{$\sigma_{trans,loc}^2$} & \textbf{0.2403} \\
\textbf{$E_{depth}$} & \textbf{0.2307} \\
$\sigma_{rot,glob}^2$ & 0.1665 \\
$\sigma_{rot,loc}^2$ & 0.1167 \\
\hline
\end{tabular}
\vspace{-7mm}
\end{table}

\section{Detailed Experimental Setup}
\subsection{Dataset Details}\label{sec:dataset}

As referenced in ~\cref{sec:details} of the main paper, our evaluation dataset consists of 1296 videos. This total includes 300 real-world sequences, sampled from nuScenes~\cite{caesar2020nuscenes}, RT-1~\cite{brohan2022rt}, and OpenVid~\cite{nan2024openvid}, which establish the ground truth for consistency. The remaining 996 generative videos are sourced from 10 diverse models, including Text-to-Video (T2V), Image-to-Video (I2V), and Video-to-Video (V2V) paradigms. A comprehensive statistical breakdown of each video source—detailing its type, task, resolution, frame rate, and the total number of videos—is provided in ~\cref{tab:genworld-stats}.

\subsection{Baseline Details}\label{sec:baseline}
\noindent\textbf{FVD.}
We followed the approach described in ~\cite{ge2024content}, using VideoMAE to extract a spatio-temporal representation consisting of 16 consecutive frames from each sequence, with a resolution of 256×256. The divergence is then calculated by comparing the mean and covariance of these generated feature embeddings against precomputed statistics from the Kinetics-400 reference set. Incorporating this strictly standardized FVD formulation ensures that our comparative analysis accurately reflects the established trade-offs between traditional fidelity-centric performance and the foundational structural stability quantified by SGC.

\begin{table*}[t]
  \centering
  \caption{Overview of video datasets used in the experiments.}

  \label{tab:genworld-stats}
  \setlength{\tabcolsep}{6pt}
  \vspace{-3mm}
  \begin{tabular}{@{}ccccccc@{}}
    \specialrule{1.2pt}{0pt}{0pt}
    \textbf{Video Source}
      & \multicolumn{1}{|c|}{\textbf{Type}}
      & \textbf{Task}
      & \textbf{Resolution}
      & \textbf{FPS}
      & \textbf{Length}
      & \multicolumn{1}{|c}{\textbf{Total}} \\
    \midrule
    
    nuScenes~\cite{caesar2020nuscenes}
      & \multicolumn{1}{|c|}{\multirow{3}{*}{Real}}
      & --   & 900--1600        & 12  & 20s
      & \multicolumn{1}{|c}{100}     \\
    RT-1~\cite{brohan2022rt}
      & \multicolumn{1}{|c|}{}
      & --   & 256--320         & 10  & 2--3s
      & \multicolumn{1}{|c}{100}    \\
    OpenVid~\cite{nan2024openvid}
      & \multicolumn{1}{|c|}{}
      & --   & 1080--1920       & 24  & 3--4s
      & \multicolumn{1}{|c}{100}    \\
    \midrule
    
    OpenSora-t~\cite{zheng2024open}
      & \multicolumn{1}{|c|}{\multirow{10}{*}{Generative}}
      & T2V  & 512$\times$512   & 8   & 2s
      & \multicolumn{1}{|c}{70}    \\
    OpenSora-i~\cite{zheng2024open}
      & \multicolumn{1}{|c|}{}
      & I2V  & 512$\times$512   & 8   & 2s
      & \multicolumn{1}{|c}{75}    \\
    Latte~\cite{ma2024latte}
      & \multicolumn{1}{|c|}{}
      & T2V  & 512$\times$512   & 8   & 2s
      & \multicolumn{1}{|c}{117}    \\
    Seine~\cite{chen2023seine}
      & \multicolumn{1}{|c|}{}
      & I2V  & 1024$\times$576  & 8   & 2--4s
      & \multicolumn{1}{|c}{108}    \\
    Zeroscope~\cite{zeroscope2024}
      & \multicolumn{1}{|c|}{}
      & T2V  & 1024$\times$576  & 8   & 3s
      & \multicolumn{1}{|c}{87}    \\
    Modelscope~\cite{wang2023modelscope}
      & \multicolumn{1}{|c|}{}
      & T2V  & 256$\times$256   & 8   & 4s
      & \multicolumn{1}{|c}{118}    \\
    VideoCrafter~\cite{chen2024videocrafter2}
      & \multicolumn{1}{|c|}{}
      & T2V  & 1024$\times$576  & 8   & 2s
      & \multicolumn{1}{|c}{125}    \\
    Hotshot~\cite{hotshot2023}
      & \multicolumn{1}{|c|}{}
      & T2V  & 672$\times$384   & 8   & 1s
      & \multicolumn{1}{|c}{105}    \\
    Lavie~\cite{wang2025lavie}
      & \multicolumn{1}{|c|}{}
      & T2V  & 1280$\times$2048 & 8   & 2s
      & \multicolumn{1}{|c}{126}    \\
    Cosmos~\cite{agarwal2025cosmos}
      & \multicolumn{1}{|c|}{}
      & V2V  & 640$\times$1024  & 8   & 1--5s
      & \multicolumn{1}{|c}{65}    \\
    \midrule
    \specialrule{1.2pt}{0pt}{0pt}
  \end{tabular}
\vspace{-7mm}
\end{table*}

\noindent\textbf{MEt3R.}
We adapt the MEt3R baseline using a output-level Moving Object Segmentation (MOS) masking strategy. Standard MEt3R inherently conflates legitimate dynamic foreground motion with background geometric instability, creating an uneven evaluation landscape when compared to SGC. Because applying artificial occlusions directly to the input frames would violate the natural image priors established during MEt3R’s training, we evaluate pristine 256$\times$256 frame pairs through the DINOv2-ViT-B/16 and MASt3R architecture and apply binary static background masks ($M_t$) strictly to the resulting spatial consistency maps ($S_t$). By computing the masked score utilizing the intersection of valid static pixels across consecutive frames—formulated as $s_t^{masked} = \text{mean}(S_t \odot (M_t \cap M_{t+1}))$—this late-masking approach rigorously isolates background geometric integrity without corrupting the underlying learned feature distributions, aligning MEt3R's evaluation scope directly with SGC.

\noindent\textbf{VBench.} VBench natively reports 9 disjoint scores with heterogeneous scales and optimization directions, which precludes direct comparison and introduces manual interpretation bias. We resolve this by applying sign inversion to inversely scaled metrics (e.g., temporal flickering) and performing sequential Z-score standardization and MinMax scaling. The dimensions are then aggregated using absolute normalized weights derived from the first principal component (PC1) of the score matrix. This objective PCA weighting naturally prioritizes the primary variance axes of generative failure, securely assigning the highest weights to subject consistency ($0.182$), background consistency ($0.148$), motion smoothness ($0.140$), and temporal flickering ($0.133$). This statistically grounded aggregation ensures VBench is evaluated rigorously and fairly alongside SGC without heuristic tuning.

\subsection{Hyperparameter Specification}\label{sec:hyper}

The SGC pipeline key computational modules, Depth Clustering and Local Pose Estimation, are governed by a set of critical hyperparameters. These parameters, which control the granularity of the scene partitioning and the robustness of the pose solver, are specified in ~\cref{tab:hyperparameters}.

\begin{table}[t]
\centering
\caption{Key Implementation Hyperparameters.}
\vspace{-3mm}
\begin{tabular}{lll}
\hline
Parameter & Module & Value \\
\hline
$k_{seg}$ & Depth Clustering & 10 \\
$S_{min}$ & Depth Clustering & 200 \\
$P_{min}$ & Local Pose Est. & 4 \\
$pnp\_inlier\_thresh$ & Local Pose Est. & 8.0 \\
$pnp\_confidence$ & Local Pose Est. & 0.99 \\
$pnp\_iterations$ & Local Pose Est. & 100 \\
\hline
\end{tabular}
\label{tab:hyperparameters}

\end{table}

\section{Additional Analysis and Ablations}
\subsection{Validation of PCA Weighting Generalizability}\label{sec:val_pca}
To validate the generalizability of the PCA-derived weights and ensure they are not overfit to our specific test set, we performed a 10-fold cross-validation on our entire 1296-video dataset. For each iteration, PCA weights were derived from the 9 training folds and used to compute scores.

\begin{table}[t]
\centering
\caption{Stability of PCA Weights Across 10 Folds.}
\label{tab:pca_stability}
\vspace{-3mm}
\begin{tabular}{lccc}
\hline
Metric & CV Mean & CV Std. & Full Data \\
 & ($w_k$) & Dev. & ($w_k$) \\
\hline
\textbf{Global Trans. Var.} & \textbf{0.2320} & \textbf{0.0337} & \textbf{0.2459} \\
\textbf{Depth Cons. Err.} & \textbf{0.2318} & \textbf{0.0158} & \textbf{0.2307} \\
\textbf{Local Trans. Var.} & \textbf{0.2273} & \textbf{0.0318} & \textbf{0.2403} \\
Global Rot. Var. & 0.1810 & 0.0345 & 0.1665 \\
Local Rot. Var. & 0.1279 & 0.0354 & 0.1167 \\
\hline
\end{tabular}
\vspace{-7mm}
\end{table}

As demonstrated in ~\cref{tab:pca_stability}, an analysis of the fold-specific PCA loadings reveals exceptional stability, with the cross-validation mean weights closely mirroring those derived from the full dataset. Crucially, the relative diagnostic importance of the components remains strictly preserved across all iterations: global translational variance ($w = 0.2320 \pm 0.0337$), depth consistency error ($w = 0.2318 \pm 0.0158$), and local translational variance ($w = 0.2273 \pm 0.0318$) consistently dominate the first principal component. The remarkably low standard deviations ($\le 0.0354$) across all five metrics demonstrate that this hierarchy of 3D spatial geometric failures is an inherent, structural property of the evaluated models rather than a sample-specific artifact. Furthermore, the near-perfect Spearman rank correlation ($\mathbf{\rho = 0.9989 \pm 0.0009}$) between $SGC$ scores computed using fold-specific weights and the final full-dataset weights empirically proves that our objective aggregation strategy is highly robust and generalizes effectively across diverse video domains.

\subsection{Sensitivity of SGC to Motion-Segmentation Quality}\label{sec:mossensitivity}
\begin{figure}[t]
    \centering
    \includegraphics[width=1\linewidth]{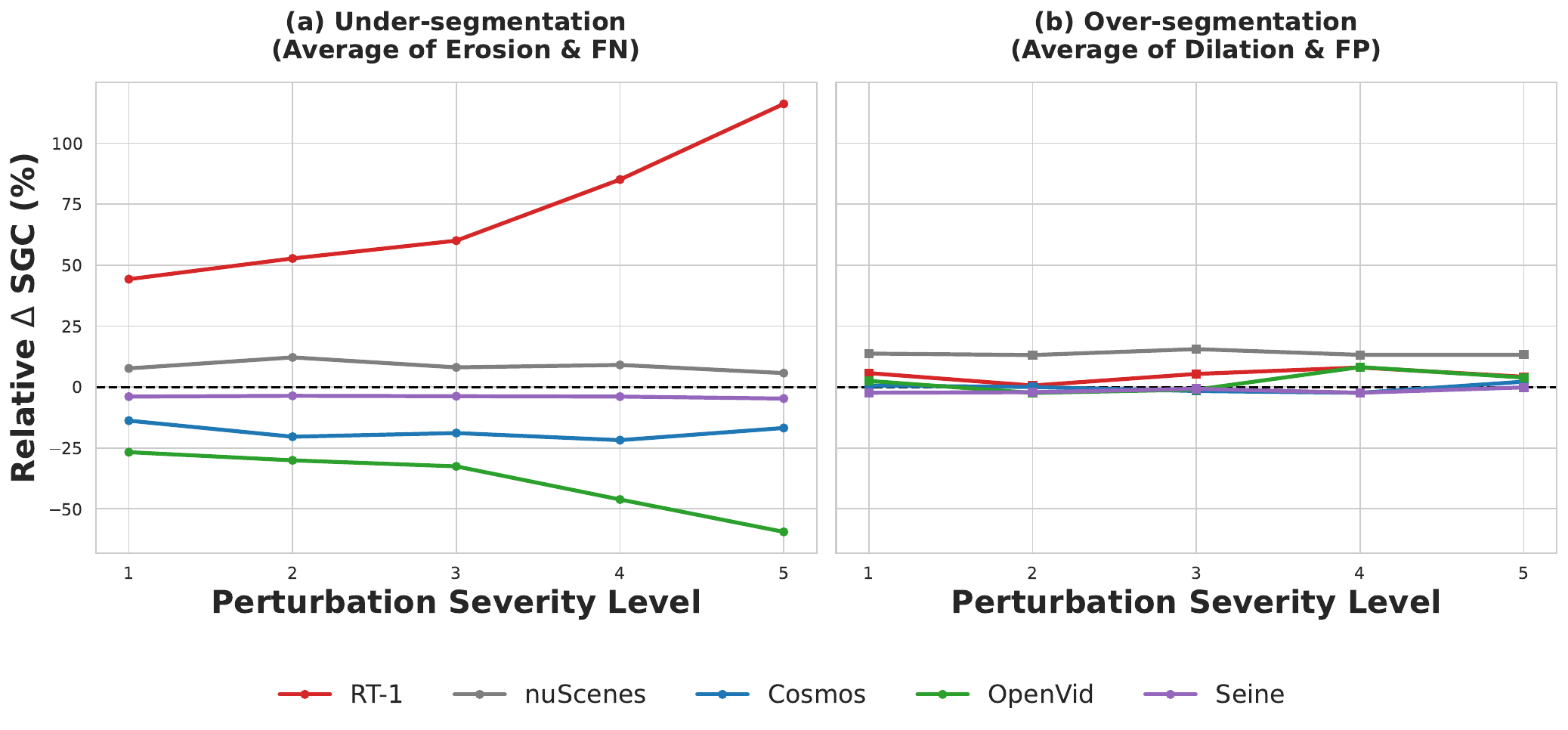}
    \vspace{-3mm}
    \caption{\textbf{Sensitivity of SGC to synthetic motion-segmentation corruption.} Relative change in SGC with respect to the clean-mask baseline under increasing under-segmentation (left) and over-segmentation (right) severity. Results are reported for three real-video datasets (nuScenes, RT-1, OpenVid) and two representative generative models (Cosmos, Seine). }
    \vspace{-7mm}
    \label{fig:sensity-mos}
\end{figure}
To rigorously evaluate SGC's sensitivity to Moving Object Segmentation (MOS) quality, we introduce synthetic mask perturbations that simulate common prediction errors. These controlled degradations consist of morphological operations and random noise injections. Morphological contraction simulates under-segmentation by producing overly conservative dynamic contours, while expansion replicates over-segmentation by encroaching upon adjacent static regions. To ensure consistent perturbation severity across varying resolutions, the magnitude of these morphological shifts scales adaptively with the image diagonal. Additionally, we inject spatially independent random noise to simulate unstructured classification failures: False Positive (FP) noise erroneously assigns dynamic labels to the background, and False Negative (FN) noise creates artificial gaps within legitimate moving targets, analogous to salt-and-pepper noise.

As illustrated in \cref{fig:sensity-mos}, SGC exhibits a predictable, scene-dependent sensitivity profile. Under-segmentation is substantially more detrimental in highly dynamic environments. For instance, the RT-1 dataset shows a relative error increase exceeding 100\% at maximum perturbation severity, confirming that unmasked foreground motion severely corrupts static background pose estimation. Conversely, the impact of over-segmentation is generally mild, with most score deviations clustering near zero. However, complex multi-object domains like nuScenes exhibit slight systematic degradation as valid background support is artificially eroded. Interestingly, for datasets like OpenVid and generative models like Cosmos, under-segmentation paradoxically reduces the SGC error, indicating that omitting challenging spatial regions can yield spurious score improvements. Ultimately, these trends demonstrate that while accurate MOS is crucial, SGC’s sensitivity is physically explainable and dictated by foreground motion intensity and intrinsic scene composition.

\begin{table}[t] 
\centering
\footnotesize
\vspace{-3mm}
\caption{\textbf{Correlation with human judgments.} SGC aligns best with human perception of 3D background stability.}
\label{tab:human_alignment}
\setlength{\tabcolsep}{12pt}
\begin{tabular}{lcc}
\toprule
\textbf{Metric} & \textbf{Spearman $\rho$} & \textbf{Kendall $\tau$} \\
\midrule
SGC & \textbf{0.945} & \textbf{0.818} \\
MEt3R~\cite{asim24met3r} & 0.864 & 0.745 \\
MEt3R(+MOS) & 0.773 & 0.600 \\
VBench-weighted & 0.664 & 0.564 \\
VBench-BC & 0.400 & 0.273 \\
\bottomrule
\end{tabular}
\vspace{-5mm}
\end{table}

\section{Alignment with Human Judgments}\label{sec:userstudy}
To rigorously validate that our metric captures true human perception of 3D spatial stability, we conducted a blinded pairwise human study. We sampled 150 video pairs across diverse generative models and real-video anchors. Seven annotators evaluated each pair with randomized A/B ordering, yielding 1,050 total judgments. Crucially, annotators were explicitly instructed to evaluate the stability of the static 3D background geometry, deliberately prioritizing geometric consistency over object motion consistency by strictly ignoring valid foreground dynamics, sharpness, and prompt alignment. We aggregated the judgments into a Human Score ($\frac{\text{Win}+0.5\times\text{Tie}}{\text{Valid}}$) and computed correlations at the 11-entity level (10 generative models and 1 merged real-video group). 

As shown in \cref{tab:human_alignment}, SGC achieves the highest correlation with human judgments (Spearman $\rho=0.945$, Kendall $\tau=0.818$), significantly outperforming 2D feature tracking approaches like MEt3R and holistic proxies like VBench-BC. This strong alignment confirms that SGC accurately quantifies the foundational 3D structural integrity that humans naturally perceive.

\section{Additional Qualitative Results}
This section provides qualitative visualizations to substantiate the statistical findings presented in the main paper. We contrast examples of high geometric consistency against specific failure modes to demonstrate the SGC metric’s diagnostic precision.

\subsection{Qualitative Validation of High Geometric Consistency}
\begin{figure}[t]
    \centering
    \includegraphics[width=1\linewidth]{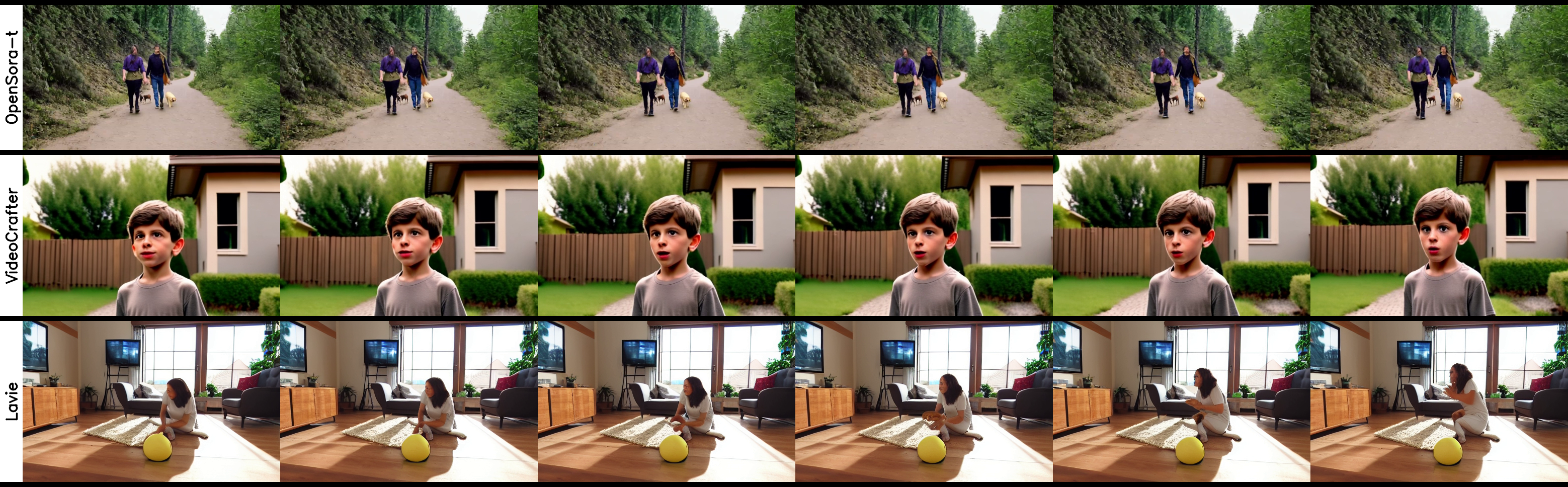}
    \vspace{-3mm}
    \caption{Qualitative Examples of High Geometric Consistency.}
    \vspace{-7mm}
    \label{fig:goodvis}
\end{figure}

~\cref{fig:goodvis} visualizes generated samples that achieve SGC scores in the top percentiles of the evaluation dataset ($\approx 0.00$), confirming the metric ability to identify physically plausible 3D environments.

Row 1 (OpenSora-t) demonstrates near-perfect scene stability. In this sequence, the SGC metric records negligible local rotation and translation variance. This quantitative result implies that all spatially distinct sub-regions of the static background maintain a rigid consensus regarding the camera trajectory, effectively mirroring the coherence expected in real-world footage.

Row 2 (VideoCrafter) highlights the metric's instance-level granularity. Although VideoCrafter exhibits variance in overall model-level performance (as seen in failure cases), this specific clip maintains remarkably low Local Translational Variance. SGC correctly identifies this instance's stability, demonstrating that the metric evaluates the geometric integrity of the specific video sample rather than biasing against the underlying model architecture.

Row 3 (Lavie) serves to validate the robustness of our Static Background Isolation module. Despite the complex dynamic foreground, the metric records a minimal depth consistency error. This confirms that the metric successfully isolates the rigid background structure (the room) from the moving subjects, ensuring that valid foreground dynamics do not penalize the consistency score.

\subsection{Diagnosis of Geometric Failure Modes}
\begin{figure*}[t]
    \centering
    \includegraphics[width=1\linewidth]{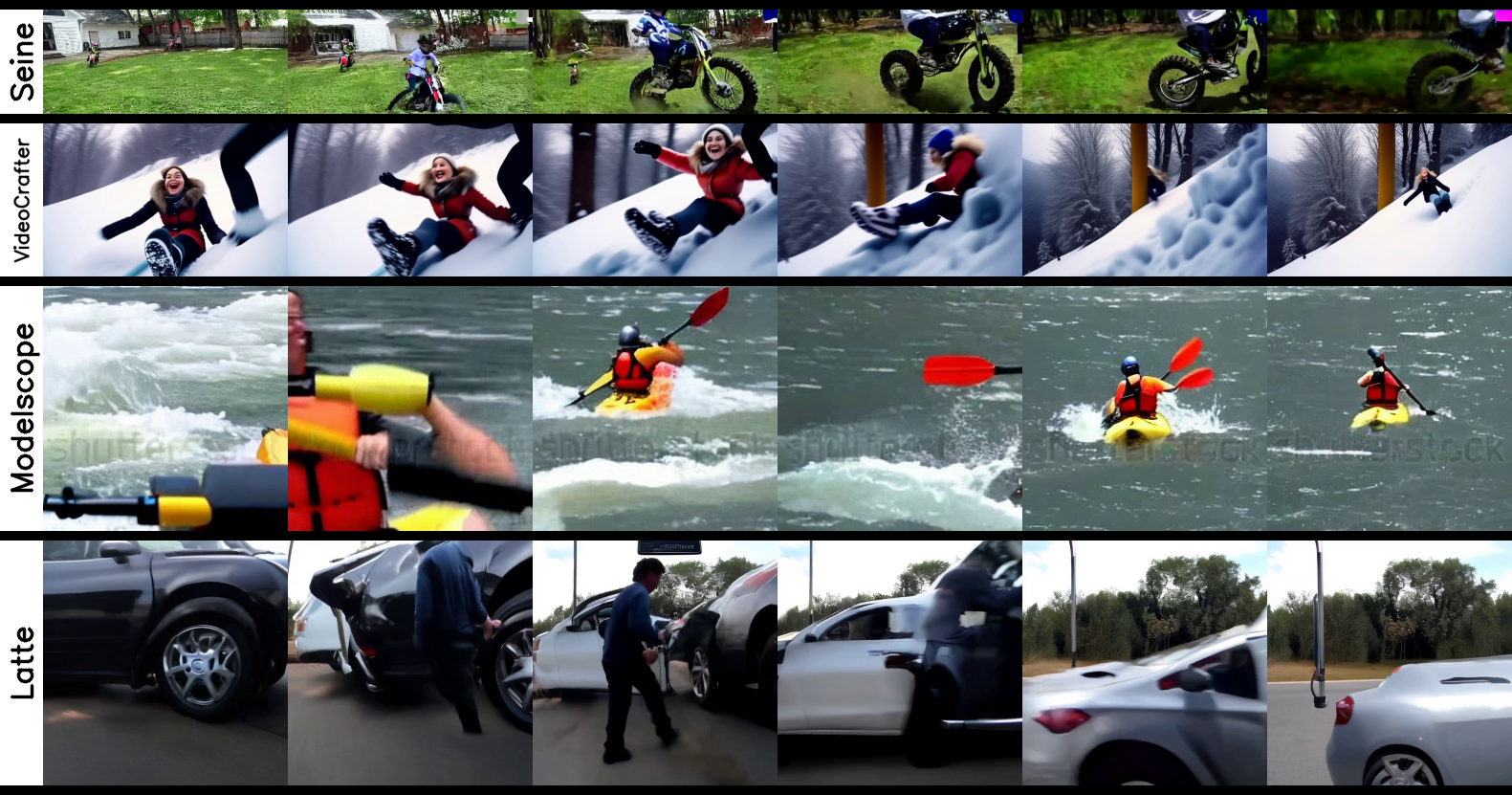}
    \vspace{-3mm}
    \caption{Diagnosis of Geometric Failure Modes.}
    \vspace{-7mm}
    \label{fig:badvis}
\end{figure*}

~\cref{fig:badvis} presents samples with high SGC scores (> 0.60) to illustrate how specific component metrics detect distinct non-physical phenomena.

Row 1 (Seine) exhibits severe Geometric Warping. The average global rotation variance is exceptionally high in this sequence. This metric spike captures the model failure to maintain a stable ground plane orientation, causing the background geometry to fluctuate wildly between frames rather than aligning with a coherent global camera trajectory.

Row 2 (VideoCrafter) illustrates a violation of 3D Perspective. Here, the Local Rotational Variance and Local Translational Variance are significantly elevated. This divergence indicates that local patches of the snowscape are shifting and warping independently of one another, failing to adhere to the rigid transformation required by the camera motion.

Row 3 (Modelscope) suffers from Object Impermanence. The depth consistency error for this clip is an order of magnitude higher than the dataset average. This metric accurately captures the non-rigid morphing of the paddle and the fluctuating water texture, signaling a failure to maintain a consistent 3D structural representation over time.

Row 4 (Latte) depicts a Catastrophic Collapse, where the car and person merge into a single amorphous structure. This breakdown results in a massive spike in Depth Consistency Error and Global Translational Variance, as the local feature tracks lose all correspondence to any stable 3D geometry, physically breaking the scene's continuity.